\documentclass[10pt,twocolumn,letterpaper]{article}

\usepackage{iccv}              

\definecolor{iccvblue}{rgb}{0.21,0.49,0.74}
\usepackage[pagebackref,breaklinks,colorlinks,allcolors=iccvblue]{hyperref}
\usepackage{xcolor}
\usepackage{pgfplots}
\usepackage{algorithm}
\usepackage{algpseudocode}
\usepackage{amsmath}
\usepackage{bbding}
\usepackage{booktabs}
\usepackage{multirow}
\usepackage{listings}
\usepackage{colortbl}
\usepackage{pifont}
\usepackage{graphicx}
\usepackage{color}
\usepackage{listings}
\usepackage{pifont} 
\usepackage{makecell}
\usepackage{longtable}
\usepackage{afterpage}

\def\paperID{10320} 
\def\confName{ICCV}
\def\confYear{2025}

\definecolor{myblue}{RGB}{233, 241, 249}
\definecolor{mygray}{RGB}{99, 110, 114}
\definecolor{myred}{RGB}{255, 118, 117}
\definecolor{myyellow}{RGB}{255, 234, 167}
\definecolor{mygreen}{RGB}{216, 226, 204}

\title{%
\begin{minipage}{0.065\textwidth} 
    \includegraphics[width=\linewidth]{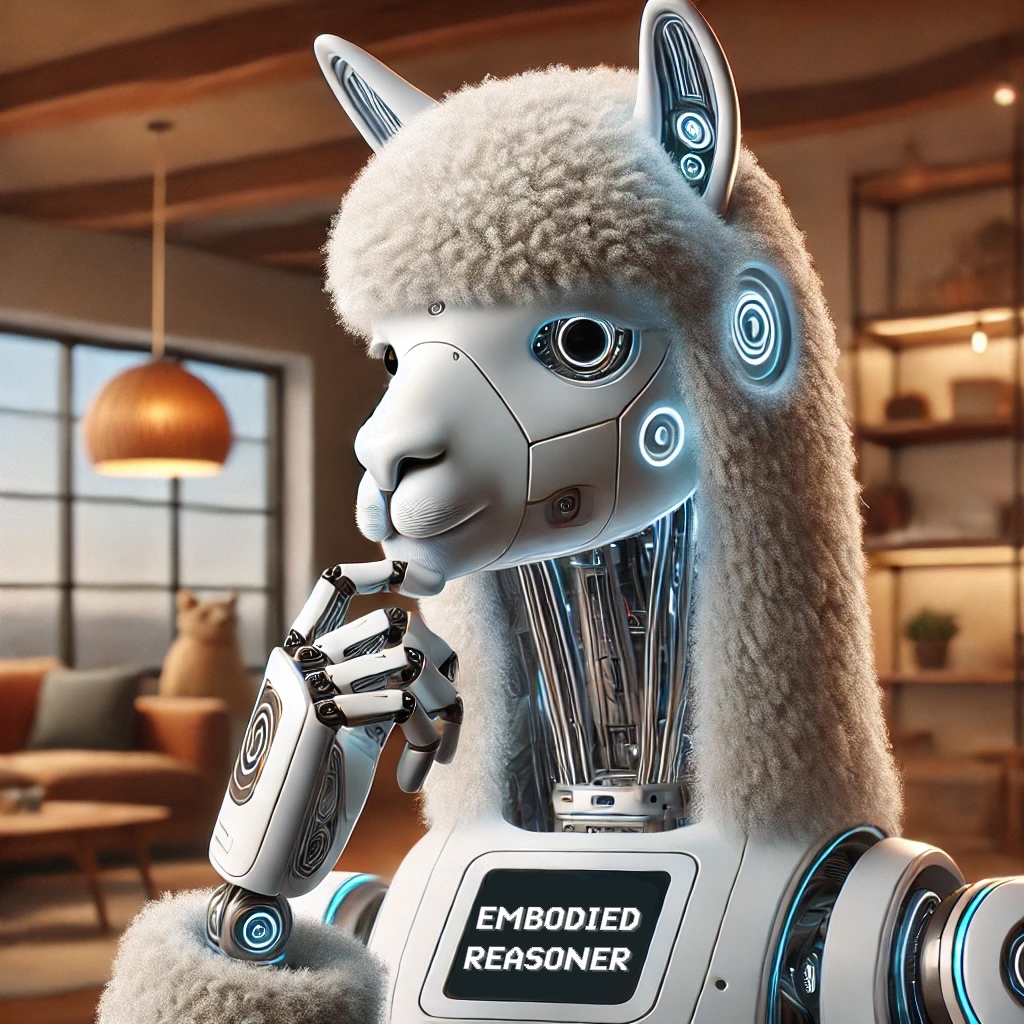}  
\end{minipage}%
\hfill
\begin{minipage}{0.91\textwidth}
    \raggedright
    {\Large \bfseries Embodied-Reasoner: Synergizing Visual Search, Reasoning, and Action\\
     for Embodied Interactive Tasks}
\end{minipage}

}

\author{Wenqi Zhang\textsuperscript{1, *}
\hspace{0.1in} Mengna Wang\textsuperscript{2,3,*}
\hspace{0.1in} Gangao Liu\textsuperscript{2,3}
\hspace{0.1in} Xu Huixin\textsuperscript{2,6,7}
\hspace{0.1in} Yiwei Jiang\textsuperscript{2,6,8}\\
\hspace{0.1in} Yongliang Shen\textsuperscript{1}  
\hspace{0.1in} Guiyang Hou\textsuperscript{1}
\hspace{0.1in} Zhe Zheng\textsuperscript{1}
\hspace{0.1in} Hang Zhang\textsuperscript{4}
\hspace{0.1in} Xin Li\textsuperscript{5}\\
\hspace{0.1in} Weiming Lu\textsuperscript{1}
\hspace{0.1in} Peng Li\textsuperscript{2,3,6,\textdagger}
\hspace{0.1in} Yueting Zhuang\textsuperscript{1,\textdagger}
\vspace{10pt}
\\
\textsuperscript{1}{College of Computer Science and Technology, Zhejiang University} \\
\textsuperscript{2}{Institute of Software, Chinese Academy of Sciences} 
\textsuperscript{3}{University of Chinese Academy of Sciences} \\
\hspace{0.3in} \textsuperscript{4}{Alibaba Group}
\textsuperscript{5}{DAMO Academy, Alibaba Group}
\textsuperscript{6}{Nanjing Institute of Software Technology} \\
\textsuperscript{7}{Nanjing University of Posts and Telecommunications} 
\textsuperscript{8}{Hohai University} \\
zhangwenqi@zju.edu.cn, lipeng@iscas.ac.cn  \\ 
\text{Project: \url{https://embodied-reasoner.github.io/}} 
}

\begin{document}
\maketitle
\renewcommand{\thefootnote}{\fnsymbol{footnote}} 
\footnotetext[1]{The first two authors have equal contributions.}  
\footnotetext[2]{Corresponding author.}  
\begin{abstract}

Recent advances in deep thinking models have demonstrated remarkable reasoning capabilities on mathematical and coding tasks. 
However, their effectiveness in embodied domains which require continuous interaction with environments through image action interleaved trajectories remains largely -unexplored. We present Embodied Reasoner, a model that extends o1 style reasoning to interactive embodied search tasks. Unlike mathematical reasoning that relies primarily on logical deduction, embodied scenarios demand spatial understanding, temporal reasoning, and ongoing self-reflection based on interaction history. To address these challenges, we synthesize 9.3k coherent Observation-Thought-Action trajectories containing 64k interactive images and 90k diverse thinking processes (analysis, spatial reasoning, reflection, planning, and verification). We develop a three-stage training pipeline that progressively enhances the model's capabilities through imitation learning, self-exploration via rejection sampling, and self-correction through reflection tuning.
The evaluation shows that our model significantly outperforms those advanced visual reasoning models, e.g., it exceeds OpenAI o1, o3-mini, and Claude-3.7 by +9\%, 24\%, and +13\%. Analysis reveals our model exhibits fewer repeated searches and logical inconsistencies, with particular advantages in complex long-horizon tasks. Real-world environments also show our superiority while exhibiting fewer repeated searches and logical inconsistency cases. 




\end{abstract}    
\section{Introduction}

\label{sec:intro}
\begin{figure}[t!]
  \centering
   \includegraphics[width=1\linewidth]{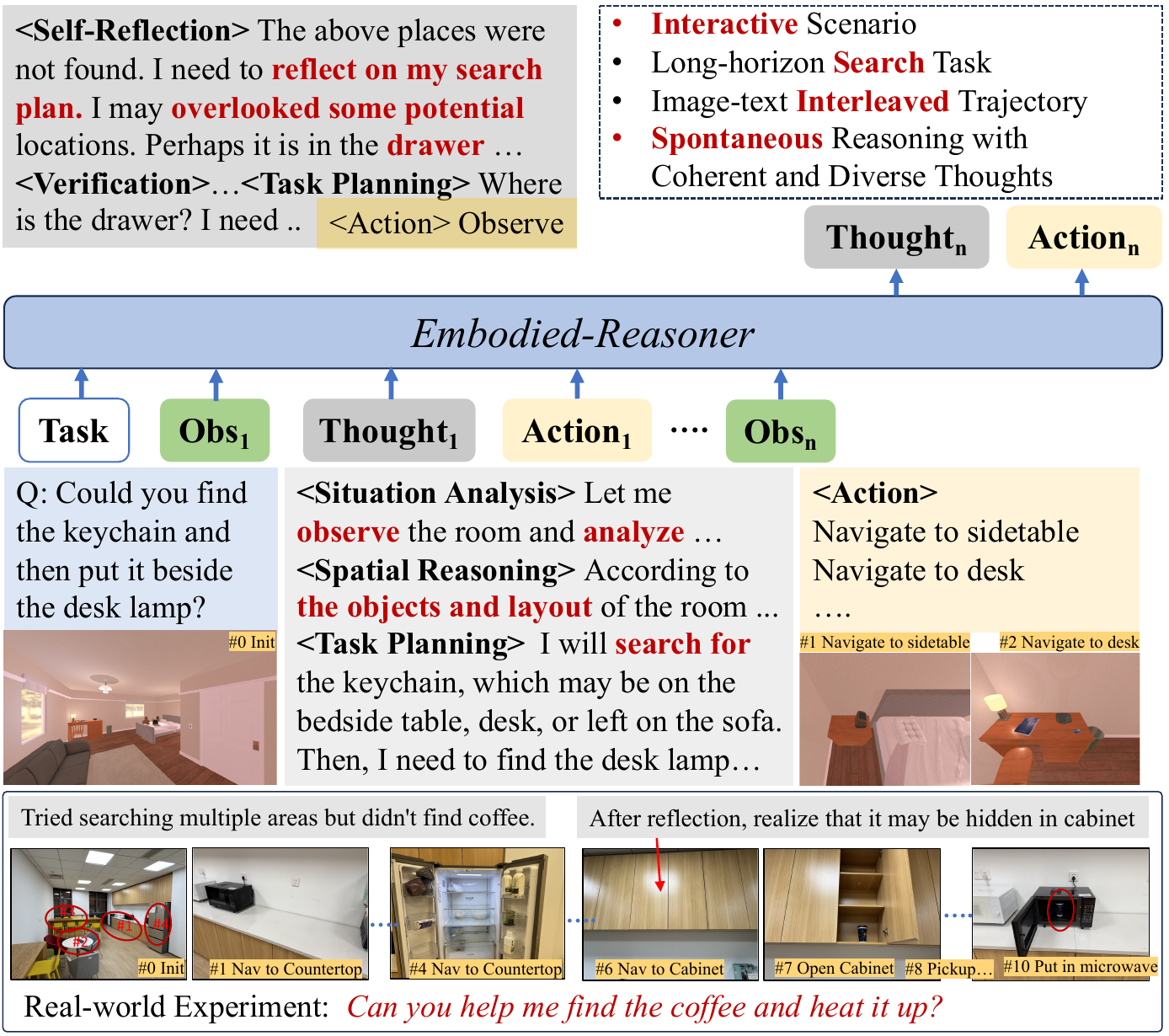}
   \caption{We design an embodied interactive task: searching for objects in an unknown room. Then we propose \emph{Embodied-Reasoner}, which presents \textbf{spontaneous reasoning and interaction} ability. Before each action, it generates diverse thoughts, e.g., self-reflection or spatial reasoning, forming an \textit{image-text} interleaved trajectory. It shows consistent reasoning and efficient search behaviors, whereas OpenAI
    o3-mini often exhibits repetitive searches and logical inconsistencies with higher failure rates.}
   \label{fig0}
\end{figure}

\begin{figure*}[h!]
  \centering
   \includegraphics[width=0.94\linewidth]{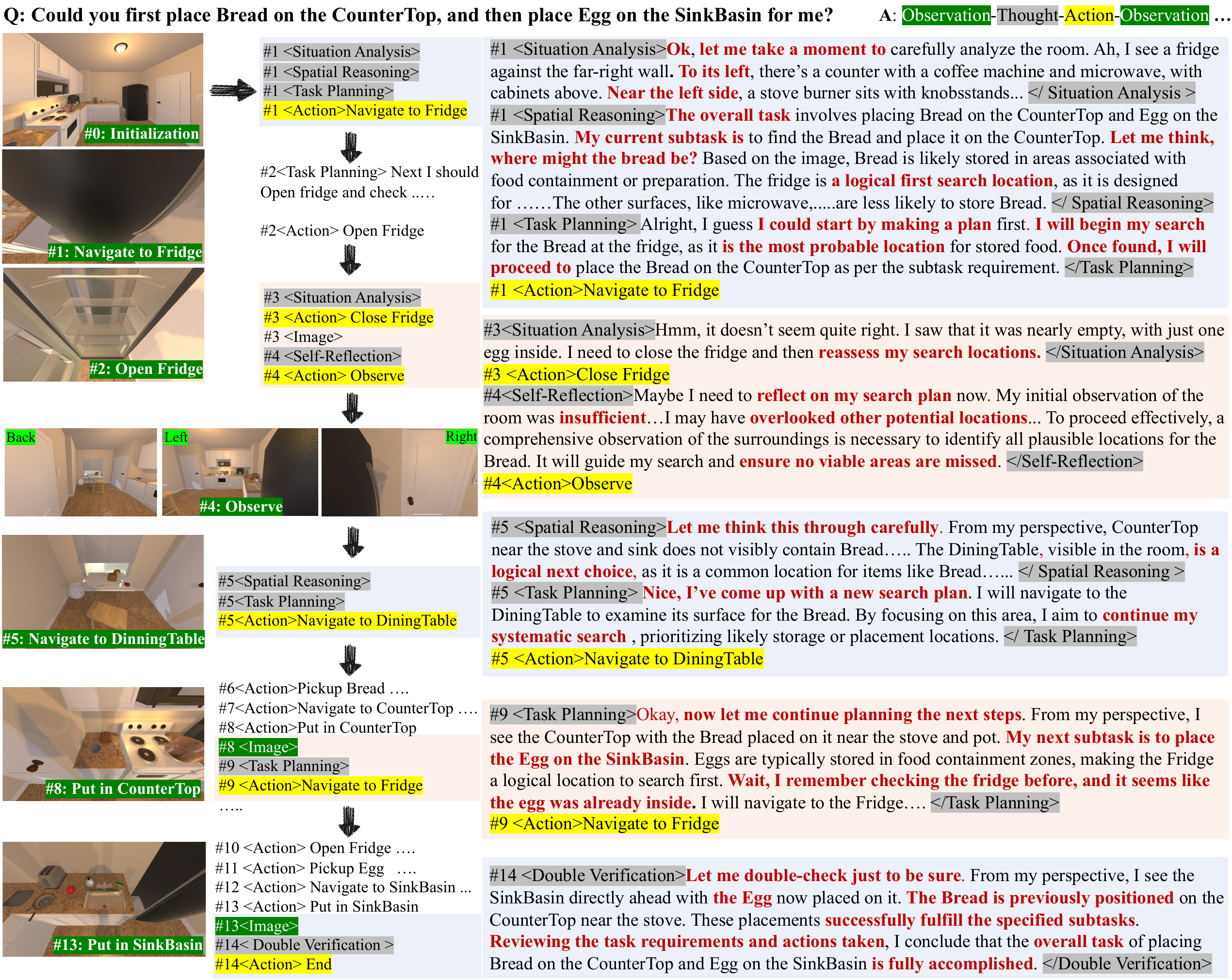}
   \caption{\textit{Embodied-Reasoner} exhibits spontaneous thinking behaviors, e.g., \emph{\textcolor{Maroon}{analyzing} environmental states} (\#1,3), \emph{\textcolor{Maroon}{reflecting} on missed details} (\#4), \emph{\textcolor{Maroon}{reasoning} based on the latest observations} (\#5), and \emph{\textcolor{Maroon}{recalling cues} for efficient planning} (\#9). These thoughts remain coherent and logically consistent despite spanning multiple rounds. In contrast, general VLMs lacking thinking abilities struggle with long-horizon interactive tasks and produce unreasonable actions, e.g., forget tasks or repetitive searching.}
   \label{fig1}
\end{figure*}

Recent advancements in deep-thinking models such as OpenAI o1~\cite{openai-o1}, Gemini 2.0 Flash Thinking~\cite{deepmind2025flashthinking}, DeepSeek R1~\cite{guo2025deepseek}, and Qwen-QwQ~\cite{qwq-32b-preview} have demonstrated remarkable reasoning capabilities in domains requiring extensive deliberation. These models, trained through large-scale reinforcement learning (RL)\cite{team2025kimi,guo2025deepseek} or post-training on elaborate thought trajectories\cite{zhao2024marco,min2024imitate}, exhibit human-like thinking patterns and self-reflection before arriving at solutions.
Their success has led to significant progress in domains requiring deliberate reasoning, particularly in college-level mathematics~\cite{guan2025rstar,min2024imitate} and coding tasks~\cite{zhang2024o1,huang2025o1}.



Despite these advances, a critical question emerges: \emph{Can the o1-style reasoning paradigm be extended beyond these specialized domains to address more complex challenges that require embodied intelligence?} Particularly, can these reasoning capabilities be effectively applied to embodied tasks demanding long-horizon planning and deliberate reasoning in interactive environments~\cite{shridhar2020alfred}? This extension is non-trivial due to several fundamental challenges:







\paragraph{Challenge 1: Extended Multimodal Interaction.} Compared to most question-answering tasks, which are limited to \textcolor{cyan}{single-turn} dialogues, embodied models operate in an \textcolor{Maroon}{interactive manner} over long-horizon task.
This means them must continuously interact with the environment, gather real-time feedback, which most appear in visual modality, and then make reasonable actions accordingly (textual modality). In such scenarios, the model need to process lengthy and image-action interleaved context and produce coherent, contextually consistent reasoning. Nevertheless, this remains a challenge for many current multimodal models and visual reasoning models~\cite{du2025virgo,guo2024mammoth,yao2024mulberry,qwenlm2025qvq,guo2024mammoth,yao2024mulberry}. We observe that even advanced reasoning models like OpenAI o3-mini~\cite{o3-mini} frequently fail to exhibit robust reasoning in these embodied interactive tasks, leading to repetitive or inconsistent behaviors.

\paragraph{Challenge 2: Diverse Reasoning Modalities.}
Different from mathematical tasks that primarily rely on \textcolor{cyan}{professional knowledge} and \textcolor{cyan}{logical deduction}, embodied scenarios demand a broader set of capabilities existing in daily life. As shown in~\cref{fig1}, when searching for an object hidden in an unknown room, the model must leverage \textcolor{Maroon}{commonsense knowledge} to infer potential search areas (e.g., \emph{steps 1, 3}), comprehend object \textcolor{Maroon}{spatial relationships} to plan efficient exploration paths at \emph{steps 1, 5}, and employ \textcolor{Maroon}{temporal reasoning} to recall relevant cues from previous attempts (\emph{step 9}) while \textcolor{Maroon}{reflecting} on prior failures. These multifaceted reasoning requirements pose challenges for multimodal models.

In this paper, we present Embodied-Reasoner, a novel approach that extends deep-thinking capabilities to embodied interactive tasks. Our key insight is that effective embodied reasoning requires not just the ability to process multimodal inputs, but also to generate diverse thinking processes (analysis, planning, reflection) that adapt to different stages of an interaction. To develop this capability, we develop a data engine that automatically synthesizes coherent \emph{Observation-Thought-Action} trajectories enriched with diverse, embodied-specific thinking processes, e.g., \emph{situational analysis, spatial reasoning, self-reflection, task planning}, and \emph{verification}. These coherent, image-text interleaved trajectories guide the model to learn how to plan and reason based on its interaction history and spatial layout, thereby boosting its spatial and temporal reasoning capabilities. We further introduce a three-stage iterative training pipeline for embodied model that combines imitation, self-exploration, and self-correction. The pipeline begins with imitation learning on synthesized trajectories to develop basic interaction skills, followed by rejection sampling tuning to enhance exploration abilities, and concludes with reflection tuning to foster self-correction.

We evaluate our approach on four high-level embodied tasks in the AI2-THOR simulator~\cite{kolve2017ai2}: Search, Manipulation, Transportation, and Composite Tasks. These tasks require agents to locate hidden objects in unfamiliar environments through reasoning and planning, then manipulate or transport them to designated areas. Our data engine synthesizes 9.3k task instructions paired with interactive trajectories, containing 64k images and 8M thought tokens, spanning 107 diverse indoor scenes, 2,100 objects, and 2,600 containers. These trajectories are used for model training.

Across 809 tasks in 12 novel scenarios, Embodied-Reasoner significantly outperforms state-of-the-art VLMs and visual reasoning models, exceeding OpenAI o1, o3-mini, and Claude-3.7-Sonnet-thinking by +9\% in success rate and +12\% in search efficiency. The performance gap widens particularly for complex composite tasks, where our model outperforms the second-best model by +39.9\%. Our analysis reveals that Embodied-Reasoner demonstrates more consistent reasoning and efficient search behaviors by spontaneously generating more reasoning tokens for complex tasks and avoiding repetitive and inefficient exploration through temporal reasoning.

Our contributions include: (1) A framework for extending deep-thinking to embodied scenarios that addresses the unique challenges of interactive reasoning; (2) A data engine that synthesizes diverse embodied reasoning trajectories with interleaved observations, thoughts, and actions; (3) A three-stage training pipeline that progressively enhances interaction, exploration, and reflection capabilities; and (4) Extensive evaluation showing significant improvements over state-of-the-art models, particularly for complex long-horizon tasks.

%

\label{sec:intro}
\begin{figure*}[t!]
  \centering
   \includegraphics[width=0.91\linewidth]{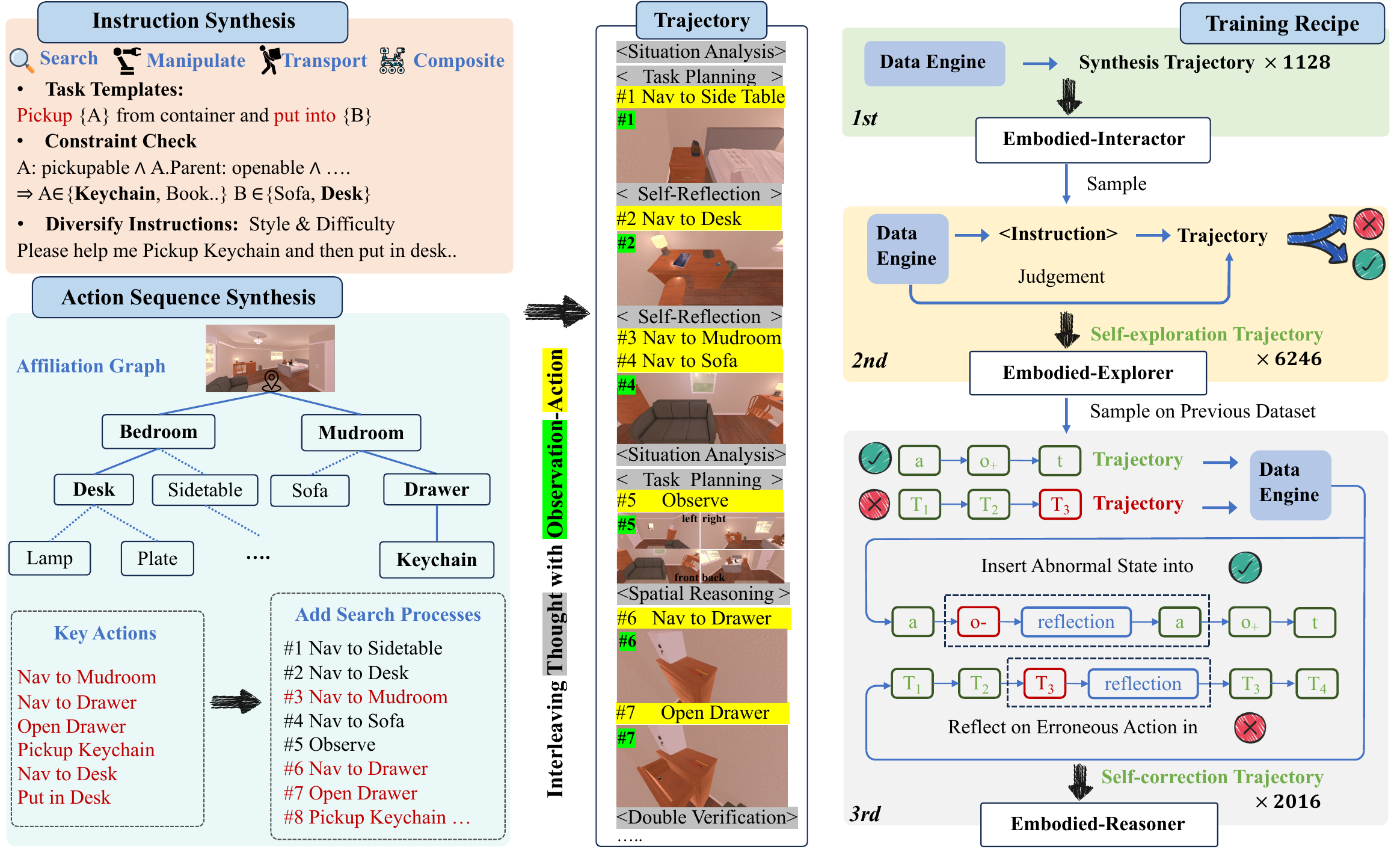}
   \caption{\emph{Left}: Data Engine for $<$Instruction, Interactive Trajectory$>$ synthesis. First, we synthesize instructions from task templates, and build an affiliation graph from scene's meta-data. It enables us to derive key actions needed for task. We add exploratory actions and insert thinking thoughts between observation and actions. \emph{Right}: Three-stage training recipe. \ding{172}We fine-tune on synthesized trajectory to develop interaction skills. \ding{173}We sample multiple trajectories on novel tasks and evaluate their correctness. The successful ones are used for developing its exploring abilities. \ding{174}We continue to sample trajectories using updated model, injecting anomalous states and reflective thoughts in successful cases and correcting errors in failed ones. This self-correction training yields \emph{Embodied-Reasoner}.}
   \label{fig_method}
\end{figure*}



\section{Observation-Thought-Action Corpora}
To develop o1-style reasoning models for embodied scenarios, we first design an embodied task that requires high-level planning and reasoning rather than low-level motor control, i.e., search for hidden objects (\cref{sec.method.task}). Next, we design a data engine in the simulator to synthesize interactive reasoning corpora: task instructions (\cref{sec.method.instruction}) and corresponding key action sequences (\cref{sec.method.action}). Each action produces a visual observation, forming an interaction trajectory. Lastly, we generate multiple thinking thoughts for each action, e.g., situation analysis, task planning, spatial reasoning, reflection, and verification, creating an interactive reasoning corpus with \textit{Observation-Thought-Action} contexts (\cref{sec.method.thought}). 

As shown in \cref{fig1}, the model requires spatial reasoning abilities to understand the layout of the kitchen and object relations, infer potential locations (fridge, dinning table) based on commonsense knowledge, systematically search unexplored areas, and adapt its plan through real-time observations while avoiding repetitive searches.


\subsection{Embodied Interactive Task}~\label{sec.method.task}
\textbf{Task Environments.} We built our embodied task environment using the widely adopted AI2-THOR simulator, which provides physics simulation and real-time vision displays. We employ 120 unique indoor scenes, e.g., kitchens, and 2,100 objects, e.g., credit card and microwave. We control the robot’s movements (e.g., moveahead) and interactions (e.g., pickup object) using AI2THOR's API, while capturing visual observations at each step.

\textbf{Task Categories.} The robot is initialized in a corner of an unknown room with a limited view, i.e., only part of the room is visible. We design four common interactive tasks in everyday life, with different complexities. \ding{172}\textbf{Search:} Searching for an object in an unknown room, e.g., keychain. It may be placed somewhere or hidden inside a container. \ding{173}\textbf{Manipulate:} Interacting with objects after searching, such as “\emph{finding a lamp and turning on the switch}”. \ding{174}\textbf{Transport:} After finding a hidden object, transport it to another location. It involves multiple search and manipulation steps. \ding{175}\textbf{Composite task:} Involving multiple transport tasks in order, e.g., \emph{“Place the egg in the microwave, and then put it on the desk after heating. After that, find ..”}.





\textbf{Action Definition.} Although AI2-THOR provides many low-level actions, our task focuses on higher-level planning and reasoning rather than movement control. Besides, low-level actions may lead to excessive interactions, so we encapsulate $9$ high-level actions built on top of atomic actions: \textit{Observe}, \textit{Move Forward}, \textit{Navigate to \{\}}, \textit{Put in \{\}}, \textit{Pickup \{\}}, \textit{Toggle \{\}}, \textit{Close \{\}}, \textit{Open \{\}}, \textit{Termination}.

\subsection{Instruction Synthesis} \label{sec.method.instruction}


Our data engine leverages LLMs to generate task instructions automatically. However, unlike previous instruction synthesis~\citep{wang2022self,shen2023taskbench}, embodied task instruction must satisfy the constraints of the scenario, i.e., avoid referencing objects that do not exist in the current scene or involving illegal actions, e.g., ``\textit{Please move the sofa to corner}'' is invalid if the scene does not contain a sofa or sofa cannot be moved. Thus, we first design multiple task templates for each task, leverage GPT-4o’s coding capabilities to automatically select objects that meet task constraints, and diversify instructions into different styles and complexities.





\textbf{Task Templates with Constraints.} We design multiple task templates for each task. \cref{fig_method} presents a transport task template: \textit{pickup \{invisible A\} and put in \{B\}}, where A denotes a hidden object with pickupable attribute, e.g., keychain, and object \textit{B} should contain containment properties, such as drawer or desk. It ensures synthesized instruction's validity. Templates and constraints are in \cref{tab:detailed task templates}.

\textbf{Code-based Object Filter.} We instruct GPT-4o to select an appropriate task template and generate code for constraint checking based on object’s metadata. It selects objects that satisfy the constraints. We fill the template with matched objects, (\textit{A}: keychain, \textit{B}: desk), and synthesize multiple instructions with different object combinations.


\textbf{Diversify Instructions}. Lastly, it automatically diversifies instructions from two levels: \ding{172} Style: We employ GPT-4o to rewrite the filled template into multiple human-style instructions, e.g., ``\textit{I can not find my keychain. Can you help me find them  and ...}''. \ding{173} Difficulty: We sequentially combine multiple simple tasks to create composite tasks.


\subsection{Action Sequence Synthesis}  \label{sec.method.action}   
Our engine automatically annotates the key actions sequences for synthesized instructions and also produces various action sequences with additional search processes.

\textbf{Affiliation Graph.} Firstly, as shown in \cref{fig_method}, we construct an affiliation graph using the simulator's metadata. In the graph, every node represents an object, and edge denotes an affiliation relation between two objects, e.g., keychain in a drawer is depicted as leaf (keychain) connected to a parent node (drawer) with a ``include'' relationship.

\textbf{Key Action Sequence.} Then we utilize the constructed affiliation graph and synthesized instruction template to derive the minimum required action sequence (key actions) for task completion. For example, “\emph{pickup the keychain and place it in desk}”, we start from the leaf node (keychain) and trace upward to its parent node (drawer) and grandparent node (mudroom). GPT-4o generates the corresponding action sequence: \emph{A1: Navigate to Mudroom}, \emph{A2: Navigate to Drawer}, \emph{A3: Open Drawer}, \emph{A4: Pickup ..}.. All key actions are indispensable for completing tasks.


\textbf{Add Additional Searching Processes.} Beyond the key action sequences, our engine also synthesizes exploratory paths by inserting additional search processes. For instance, as shown in \cref{fig_method}, our engine first inserts three searching actions: \emph{Nav to Sidetable}, \emph{Desk}, and \emph{Sofa}. After failing to find the keychain, it inserts an \emph{Observe} action until it ultimately locates the keychain in the drawer. These additional search actions make the trajectory more realistic and reasonable, showcasing how robot gradually explores an unfamiliar environment until it successfully locates the target.





\subsection{Interleaving Thought with Observation-Action} \label{sec.method.thought}   
After running synthesized actions ($a_1,a_2,...a_n$), we obtain an interaction trajectory: $o_1, a_1, o_2, a_2, ..., o_n, a_n$, where $o_i$ denotes first-person perspective images. Then, we generate multiple deep-thinking thoughts ($t_i$) for each action, creating an interleaved context: \textit{Observation-Thought-Action}.




\textbf{Diverse Thinking Pattern.} Firstly, we define five thinking patterns to simulate human cognitive activities in different situations: \emph{Situation Analysis, Task Planning, Spatial Reasoning, Self-Reflection, and Double Verification}. We use concise prompts to describe each pattern, guiding GPT-4o in synthesizing the corresponding thinking process.

\textbf{Derive Thought from Observation-Action.} For each interaction, we instruct GPT-4o to select one or more thinking patterns and then generate detailed thoughts based on interactive context. These thoughts are inserted between observations and actions ($o_n,a_n \!\!\rightarrow\!\! o_n,t^{1}_n,t^{2}_n,..t^{k}_n,a_n$). Specifically, we prompt GPT-4o with the previous interaction trajectory ($o_1, t_1, a_1, …, o_n$), and upcoming action ($a_n$), and generate a well-reasoned thinking process ($t_n$). It should consider the latest observation ($o_{n}$) and offer reasonable rationales for next-step action ($a_n$), and also remain logically consistent with the previous thoughts ($t_{1:n\textbf{-}1}$).



\begin{figure}[t!]
  \centering
   \includegraphics[width=0.9\linewidth]{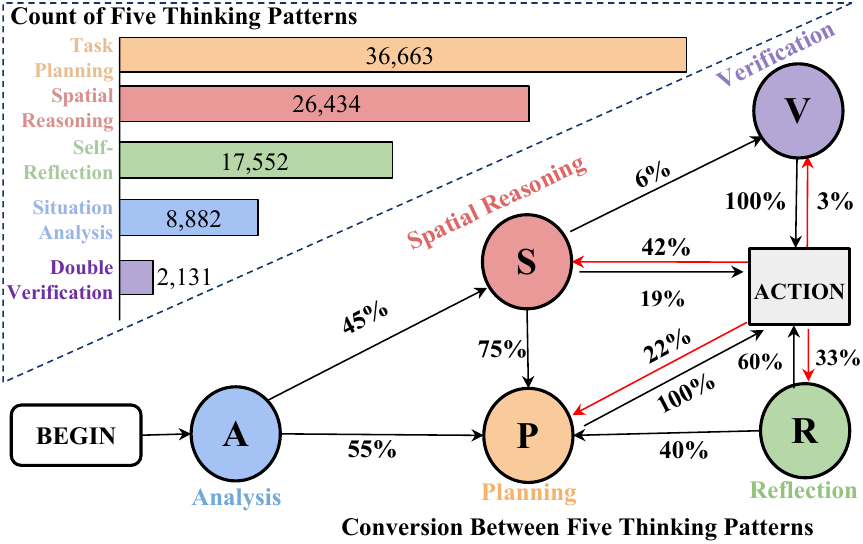}
   \caption{We analyze the frequency of five types of thoughts and their flexible transition relationships in all trajectories.}
   \label{thought_analysis}
\end{figure}

\section{Training Recipe for Embodied-Reasoner}
To incentivize the reasoning ability, we design three training stages, i.e., imitation learning, rejection sampling tuning, and reflection tuning, that bootstrap a general VLM into an embodied interactive model with deep thinking ability.

\textbf{Multi-turn Dialogue Format.} Considering that the interactive trajectory follows an interleaved image-text format (\textit{Observation-Thought-Action}), we organize them as multi-turn dialogue corpora. In each turn, the observed images and the simulator’s feedback serve as the \textit{User Input}, while thoughts and actions as \textit{Assistant Output}. During training, we compute the loss only for thoughts and action tokens. 

\subsection{Learn to Interact: Imitation Learning} 
In the first stage, we use a data engine to generate a small set of instruction-trajectory, most contain limited searching processes or only consist of key actions (\emph{Observation-Thought-Key action}). Qwen2-VL-7B-Instruct is fine-tuned on this dataset and learns to understand interleaved image-text context, output reasoning and action tokens.

After tuning, we develop \textbf{\textit{Embodied-Interactor}}, which is capable of interaction within embodied scenarios. However, most synthesized trajectories only include key actions for task completion, without searching processes or observing environments. In most cases, \textit{Embodied-Interactor} exhibits limited search capabilities, i.e., it does not know how to handle situations where objects cannot be directly found and require further searching. For instance, when it opens a refrigerator for an egg but is empty, it may respond: ``egg does not exist" rather than searching for other locations.




\subsection{Learn to Search: Rejection Sampling Tuning} 

\textbf{Self-exploration Trajectory.} DeepSeek-R1 reveals that advanced reasoning ability can be acquired through rejection sampling and reward-guided RL on large-scale self-exploration data. Inspired by this, we employ \textit{Embodied-Interactor} to sample massive self-generated trajectories for further training. Specifically, as shown in \cref{fig_method}, we employ data engine to synthesize new task instructions and their key actions, and then use \textit{Embodied-Interactor} to sample multiple trajectories on each instruction under high-temperature settings. Lastly, we select high-quality trajectories.

\textbf{Data Engine as the reward model.} We use our data engine as a process supervision reward model (PRM) to assess these sampled trajectories. We retain 6,246 successful trajectories—most of which complete the task after several search attempts. We perform second-stage instruction-tuning on all collected trajectories, developing \textit{Embodied-Explorer}. We observe it exhibits adaptive planning and searching behaviors. For example, when the target object cannot be found directly, it formulates a detailed search plan involving multiple potential areas with different priorities.


\subsection{Learn to Self-reflect: Reflection Tuning}

\textit{Embodied-Explorer} occasionally produces unreasonable actions, especially in long-horizon tasks, such as hallucination. Besides, robots often encounter temporary hardware malfunctions. It requires the model to self-reflect on unreasonable behaviors, recognize abnormal states, and correct them on time. As shown in \cref{fig_method}, we employ \textit{Embodied-Explorer} to sample massive trajectories on previous tasks. \ding{172} For failed trajectories, we locate the first erroneous actions and construct self-correction trajectories. \ding{173} For succeeded trajectories, we insert anomaly states to simulate hardware fault.

\textbf{Insert Abnormal State into Succeeded Trajectory.} We simulate two robot anomalies: \textbf{navigation anomaly}, where the robot navigates to an inconsistent location with command (e.g., action: ``\emph{navigate to the refrigerator}'', but instead navigates to table); and \textbf{manipulation anomaly}, where robot arm temporarily fails to perform interaction command. For a succeeded trajectory \{.., \textcolor{ForestGreen}{$a$}, \textcolor{ForestGreen}{$o_{+}$}, \textcolor{ForestGreen}{$t$}..\}, we insert an abnormal state (\textcolor{red}{$o_{-}$}) after an action (\textcolor{ForestGreen}{$a$}), and then generate self-reflective thoughts (\textcolor{blue}{$t_r$}) for this anomalies. Lastly, we retry the same action: \{.., \textcolor{ForestGreen}{$a$}, \textcolor{red}{$o_{-}$}, \textcolor{blue}{$t_r$}, \textcolor{ForestGreen}{$a$}, \textcolor{ForestGreen}{$o_{+}$} ..\}.

\textbf{Reflect on Unreasonable Action in Failed Trajectory.} Using synthesized key actions, we identify the first incorrect action in each failed trajectory (\textcolor{red}{$Traj_{-}$}). Then we generate self-reflective thoughts (\textcolor{blue}{$t_r$}) for incorrect action and supplement remaining correct trajectories (\textcolor{ForestGreen}{$Traj^{t:n}_{+}$}), creating a revised trajectory: \{\textcolor{red}{$Traj^{1:t}_{-}$}, \textcolor{blue}{$t^{t}_r$}, \textcolor{ForestGreen}{$Traj^{t:n}_{+}$}\}. We fine-tune the model on synthesized self-correction trajectories. For loss calculation, we mask out erroneous partial trajectory (\textcolor{red}{$Traj^{1:t}_e$}) and only compute the loss for reflective tokens (\textcolor{blue}{$t^{t}_r$}) and the correct trajectory (\textcolor{ForestGreen}{$Traj^{t:n}_c$}).






\begin{table}[t!]
    \centering 
    \footnotesize
    \setlength\tabcolsep{1pt} 
    \begin{tabular}{l|clccc}
        \toprule[1pt]
        \textbf{Stage} & \#\textbf{Trajectory} & \textbf{Source} & \makecell[l]{\#\textbf{Image}$_{all }$}& \makecell[l]{\#\textbf{Image}$_{max}$}&\makecell[l]{\#\textbf{Action}$_{avg}$}\\ 
        \toprule[0.5pt]
        Train$_{1st}$ & 1,128&\makecell[l]{Synthesis}& 4636 &11&4.11\\ 
        Train$_{2nd}$ & 6,246&\makecell[l]{Self-Explore}& 45.8k &26&7.33\\
        Train$_{3rd}$ & 2,016&\makecell[l]{Synthesis}& 13.8k&29&8.63\\  
        Total     & 9,390&- & 64k&29&7.22\\ \hline       
        Testset & 809 &Human & 4.9k&29&6.06\\
        \bottomrule[1pt]
    \end{tabular}    
    \caption{We synthesize 9.3$k$ $\langle$task, trajectory$\rangle$ for training. Also, we manually annotate key actions for 809 novel testing tasks.}
    \label{Dataset}
\end{table}

\section{Dataset Statistics}  
\subsection{Training Corpus}
As shown in~\cref{Dataset}, we synthesize 9,390 unique task instructions and their \textit{Observation-Thought-Action} trajectory for three training stages, i.e., $\langle$\textit{Scene,Inst,Traj}$\rangle$. In the first stage, data engine synthesizes 1,128 instruction-trajectory pairs. In the 2nd stage, we remain 6,246 exploratory trajectories through rejection sampling. In the 3rd stage, data engine synthesizes 2,016 self-correction trajectories.   

Our dataset spans 107 diverse indoor scenes, e.g., kitchens and living rooms, and covers 2,100 interactive objects (e.g., eggs, laptops) and 2,600 containers (e.g., refrigerators, drawers). All trajectories contain \textit{64K} a first-person perspective image from interaction and \textit{8M} thought tokens. The distribution of our dataset is presented in \cref{sec:dataset-details}.


\subsection{Thoughts Analysis}




\textbf{The Distribution of Thinking Patterns.} We count the frequency of five thinking patterns in all trajectories. As shown in \cref{thought_analysis}, \emph{task planning} and \emph{spatial reasoning} appear most frequently, with 36.6K and 26.4K, respectively. It means each trajectory contains about four \emph{planning} and three \emph{reasoning}. Besides, \emph{Self-reflection} often occurs after a search fails, with about two times per trajectory. These diverse thoughts incentivize the model’s reasoning capabilities.

\textbf{Conversion between Thinking Patterns.} We also compute the transition probabilities between five thinking patterns (see \cref{thought_analysis}). We find the relationship between them is flexible and depends on the situation. It typically begins with \textit{Task Planning}, followed by \emph{task planning} (55\%) and \emph{spatial reasoning} (45\%). When navigating unknown regions, it frequently relies on \textit{spatial reasoning} (Action$\rightarrow$S: 42\%). If a search attempt fails, it shifts to \textit{self-reflection} (Action$\rightarrow$R: 33\%), and once a (sub-)task is completed, it may perform \textit{double verification} sometimes (Action$\rightarrow$V: 3\%, S$\rightarrow$V: 6\%). This diverse structure enables the model to learn spontaneous thinking and flexible adaptability.


\subsection{Interactive Evaluation Framework}
We cultivate 809 test cases across 12 novel scenarios, which are different from training scenes. We manually design instructions and annotate corresponding key actions and final states: $\langle$Instruction, Key Action, Final state$\rangle$. Notably, our test-set contains 25 carefully designed ultra long-horizon tasks, each involving four sub-tasks and 14-27 key actions.

\textbf{Metrics} We design three metrics to assess the quality of the model-generated trajectories. \textcolor{Maroon}{Success Rate (\%)}: It measures whether a task is successfully completed by evaluating if the key actions align correctly and if the final state meets the task criteria. \textcolor{Maroon}{Search Efficiency:} It evaluates task efficiency—more steps indicate lower efficiency. We calculate it as the ratio of key action numbers to predicted action numbers. \textcolor{Maroon}{Task Completeness (\%)}: It computes the proportion of predicted actions that belong to the set of key actions.

\begin{table*}[t]
\centering
\footnotesize
\setlength\tabcolsep{5pt} 
\begin{tabular}{lcccccccc}
\Xhline{2\arrayrulewidth}
\multicolumn{2}{c|}{\multirow{2}{*}{\textbf{Model}}} & 
\multicolumn{1}{c|}{\multirow{2}{*}{\begin{tabular}[c]{@{}c@{}}\textbf{Success}\\ \textbf{Rate} $\uparrow$ \end{tabular}}} &
\multicolumn{1}{c|}{\multirow{2}{*}{\begin{tabular}[c]{@{}c@{}}\textbf{Search}\\ \textbf{Efficiency}  $\uparrow$\end{tabular}}} &
\multicolumn{1}{c|}{\multirow{2}{*}{\begin{tabular}[c]{@{}c@{}}\textbf{Task}\\ \textbf{Completeness} $\uparrow$\end{tabular}}} &
\multicolumn{4}{c}{\textbf{Success Rate for SubTasks} $\uparrow$}  \\\cline{6-9}

\multicolumn{2}{l|}{} &\multicolumn{1}{l|}{} &\multicolumn{1}{l|}{} & \multicolumn{1}{l|}{}& \multicolumn{1}{c|}{\text{Search }} 
& \multicolumn{1}{c|}{\text{Manipulate }} & 
\multicolumn{1}{c|}{\text{Transport }} & 
\multicolumn{1}{c}{\text{Composite }}  \\ 

\Xhline{2\arrayrulewidth}


\multicolumn{1}{l|}{} & \multicolumn{1}{l|}{\cellcolor{myred!10} Qwen2.5-VL-7B-Instruct \cite{bai2025qwen2}} & \cellcolor{myred!10} 12.38\% & \cellcolor{myred!10} 10.87\% & \cellcolor{myred!10} 27.53\% \cellcolor{myred!10} & \cellcolor{myred!10} 6.45\% & \cellcolor{myred!10} 23.55\% & \cellcolor{myred!10} 7.56\% & \cellcolor{myred!10} 0.95\% \\

\multicolumn{1}{l|}{} & \multicolumn{1}{l|}{\cellcolor{myred!10} Qwen2-VL-7B-Instruct \cite{wang2024qwen2}} & \cellcolor{myred!10} 14.79\% & \cellcolor{myred!10} 11.97\% & \cellcolor{myred!10} 38.67\% & \cellcolor{myred!10} 23.33\% & \cellcolor{myred!10} 25.50\% & \cellcolor{myred!10} 2.82\% & \cellcolor{myred!10} 0.00\% \\

\multicolumn{1}{l|}{\textit{General-}} & \multicolumn{1}{l|}{\cellcolor{myred!10} Qwen2.5-VL-72B-Instruct \cite{bai2025qwen2}} & \cellcolor{myred!10} 31.75\% & \cellcolor{myred!10} 22.61\% & \cellcolor{myred!10} 50.62\% & \cellcolor{myred!10} 52.14\% & \cellcolor{myred!10} 38.89\% & \cellcolor{myred!10} 21.90\% & \cellcolor{myred!10} 0.00\% \\

\multicolumn{1}{l|}{\textit{purpose}} & \multicolumn{1}{l|}{\cellcolor{myred!10} Qwen2-VL-72B-Instruct \cite{wang2024qwen2}} & \cellcolor{myred!10} 39.00\% & \cellcolor{myred!10} 28.88\% & \cellcolor{myred!10} 54.56\% & \cellcolor{myred!10} 50.00\% & \cellcolor{myred!10} 52.36\% & \cellcolor{myred!10} 33.19\% & \cellcolor{myred!10} 0.00\% \\

\multicolumn{1}{l|}{\textit{VLMs}} & \multicolumn{1}{l|}{\cellcolor{myred!10} Claude 3.5-Sonnet \cite{claude-3.5-sonnet}} & \cellcolor{myred!10} 45.35\% & \cellcolor{myred!10} 28.05\% & \cellcolor{myred!10} 64.12\% & \cellcolor{myred!10} 54.25\% & \cellcolor{myred!10} 50.51\% & \cellcolor{myred!10} 51.22\% & \cellcolor{myred!10} 3.84\%  \\

\multicolumn{1}{l|}{} & \multicolumn{1}{l|}{\cellcolor{myred!10} Qwen-VL-Max \cite{qwen-vl-max}} & \cellcolor{myred!10} 49.81\% & \cellcolor{myred!10} 36.28\% & \cellcolor{myred!10} 68.39\% & \cellcolor{myred!10} 63.87\% & \cellcolor{myred!10} 63.21\% & \cellcolor{myred!10} 45.16\% & \cellcolor{myred!10} 1.90\% \\

\multicolumn{1}{l|}{} & \multicolumn{1}{l|}{\cellcolor{myred!10} GPT-4o \cite{gpt-4o}} & \cellcolor{myred!10} 66.67\% & \cellcolor{myred!10} 41.68\% & \cellcolor{myred!10} 79.07\% & \cellcolor{myred!10} 69.03\% & \cellcolor{myred!10} 79.26\% & \cellcolor{myred!10} 71.95\% & \cellcolor{myred!10} 14.42\%  \\

\Xhline{2\arrayrulewidth}



\multicolumn{1}{l|}{} & \multicolumn{1}{l|}{\cellcolor{myyellow!30} QVQ-72B-Preview \cite{qwenlm2025qvq}}  & \cellcolor{myyellow!30} 7.54\% & \cellcolor{myyellow!30} 6.39\% & \cellcolor{myyellow!30} 36.33\% & \cellcolor{myyellow!30} 4.35\% & \cellcolor{myyellow!30} 7.50\% & \cellcolor{myyellow!30} 10.53\% & \cellcolor{myyellow!30} 0.00\%\\

\multicolumn{1}{l|}{} & \multicolumn{1}{l|}{\cellcolor{myyellow!30} Kimi-K1.5${}^{\dagger}$  \cite{team2025kimi}} & \cellcolor{myyellow!30} 46.00\% & \cellcolor{myyellow!30} - &  \cellcolor{myyellow!30} -  & \cellcolor{myyellow!30} - & \cellcolor{myyellow!30} \cellcolor{myyellow!30} - & \cellcolor{myyellow!30} - & \cellcolor{myyellow!30} -\\

\multicolumn{1}{l|}{} & \multicolumn{1}{l|}{\cellcolor{myyellow!30} GPT-o3-mini \cite{o3-mini}} & \cellcolor{myyellow!30} 56.55\% & \cellcolor{myyellow!30} 26.93\% & \cellcolor{myyellow!30} 67.41\% & \cellcolor{myyellow!30} \textbf{78.57\%} & \cellcolor{myyellow!30} 59.32\% & \cellcolor{myyellow!30} 66.67\% & \cellcolor{myyellow!30} 0.00\% \\

\multicolumn{1}{l|}{\textit{Visual}} & \multicolumn{1}{l|}{\cellcolor{myyellow!30} Gemini-2.0 Flash Thinking \cite{deepmind2025flashthinking}}  & \cellcolor{myyellow!30} 56.74\% & \cellcolor{myyellow!30} 43.01\% & \cellcolor{myyellow!30} 71.70\% & \cellcolor{myyellow!30} 71.05\% & \cellcolor{myyellow!30} 75.60\% & \cellcolor{myyellow!30} 40.67\% & \cellcolor{myyellow!30} 8.89\% \\

\multicolumn{1}{l|}{\textit{Reasoning}} & \multicolumn{1}{l|}{\cellcolor{myyellow!30} Claude-3.7-Sonnet-thinking \cite{claude-3.7-sonnet-thinking}}  & \cellcolor{myyellow!30} 67.70\% & \cellcolor{myyellow!30} 37.95\% & \cellcolor{myyellow!30} 78.63\% & \cellcolor{myyellow!30} 69.12\% & \cellcolor{myyellow!30} 75.88\% & \cellcolor{myyellow!30} 71.94\% & \cellcolor{myyellow!30} 13.79\%\\

\multicolumn{1}{l|}{\textit{Models}} & \multicolumn{1}{l|}{\cellcolor{myyellow!30} GPT-o1 \cite{openai-o1}}  & \cellcolor{myyellow!30} 71.73\% & \cellcolor{myyellow!30} 43.06\% & \cellcolor{myyellow!30} \cellcolor{myyellow!30} 82.49\% & \cellcolor{myyellow!30} 78.42\% & \cellcolor{myyellow!30} 79.10\% & \cellcolor{myyellow!30} 67.36\% & \cellcolor{myyellow!30} 13.16\% \\

\cline{2-9}



\multicolumn{1}{l|}{} & \multicolumn{1}{l|}{\cellcolor{mygreen!50} Embodied-Interactor-7B (ours-1st)} &  
\cellcolor{mygreen!50} 25.46\% & \cellcolor{mygreen!50} 24.75\% & \cellcolor{mygreen!50} 53.67\% & \cellcolor{mygreen!50} 30.97\% & \cellcolor{mygreen!50} 27.09\% & \cellcolor{mygreen!50} 29.20\% & \cellcolor{mygreen!50} 3.81\% 
\\

\multicolumn{1}{l|}{} & \multicolumn{1}{l|}{\cellcolor{mygreen!50} Embodied-Explorer-7B (ours-2nd)} & 
\cellcolor{mygreen!50} 65.39\% & \cellcolor{mygreen!50} 46.25\% & \cellcolor{mygreen!50} 77.73\% & \cellcolor{mygreen!50} 60.00\% & \cellcolor{mygreen!50} 75.92\% & \cellcolor{mygreen!50} 72.24\% & \cellcolor{mygreen!50} 26.67\% 
\\

\multicolumn{1}{l|}{} & \multicolumn{1}{l|}{\cellcolor{mygreen!50} Embodied-Reasoner-7B (ours-3rd)} & 
\cellcolor{mygreen!50} \textbf{80.96\%} & \cellcolor{mygreen!50} \textbf{55.07\%} & \cellcolor{mygreen!50} \cellcolor{mygreen!50} \textbf{86.30\%} & \cellcolor{mygreen!50} 65.16\% & \cellcolor{mygreen!50} \textbf{93.31\%} & \cellcolor{mygreen!50} \textbf{87.20\%} & \cellcolor{mygreen!50} \textbf{54.29\%}
\\
\Xhline{2\arrayrulewidth}
\end{tabular}

\caption{We compare the performance of \emph{Embodied-Reasoner} against advanced VLMs and visual reasoning models. After the three-stage training process, we boost Qwen2-VL-7B from 14.8 to 81. Kimi-K1.5${}^{\dagger}$ means we manually evaluate 50 testing cases through the webpage.}
\vspace{-7pt}
\label{tab:main_result}
\end{table*}

        
            
        

\section{Experiments}
\subsection{Main Results}
\textbf{Much higher success rate, search efficiency, and task completeness.} As shown in ~\cref{tab:main_result}, \emph{Embodied-Reasoner} significantly outperforms all reasoning models and VLMs by a large margin (+9.6\% to GPT-o1), including the latest GPT-o3-mini (+24\%) and Claude-3.7-Sonnet-thinking (+13\%). Besides success rate, our model also demonstrates clear advantages on search efficiency and task completeness over others, e.g., search efficiency is higher than GPT-o1 by +12\%. Despite being significantly smaller than advanced reasoning models, \emph{Embodied-Reasoner} demonstrates stronger interaction and reasoning capabilities in embodied scenarios.


\textbf{Advantages are more obvious on complex tasks.} Analyzing success rate across four different task categories (search, manipulate, transport, and composite tasks), we observe that \emph{Embodied-Reasoner} demonstrates significantly stronger performance on the more challenging tasks like composite and transport. Notably, on composite tasks, it outperforms the second-best model (GPT-4o) by +39.9\%. Interestingly, on the relatively simpler search task, our model lags behind GPT-o3-mini by 13.4\%. Our analysis finds that on these simpler tasks, \emph{Embodied-Reasoner} sometimes over-explores, leading to missed detections of nearby objects.

\textbf{Our three-stage training progressively incentivize interaction and reasoning capabilities, from 14.7\% to 80.9\%.} Our base model, Qwen2-VL-7B, initially achieved only 14.7\%. After the first-stage imitation learning, it improved to 25.4\%, mastering simple interaction ability. Subsequently, rejection sampling tuning significantly boosted performance to 65.4\%, reaching a level comparable to GPT-o1 with exploration abilities. Finally, fine-tuning with self-correction trajectories further elevated the model’s success rate to 80.9\%. We observe that most baseline models often exhibit \textbf{repetitive searching} behaviors and \textbf{unreasonable planning}, especially when handling long-horizon tasks. In contrast, \emph{Embodied-Reasoner} performs a strategic search and planning after deep thinking and timely self-reflection, significantly reducing those unreasonable cases.



\subsection{Analysis: How does deep-thinking paradigm enhance embodied search tasks?}

\begin{figure}[t!]
  \centering
   \includegraphics[width=1\linewidth]{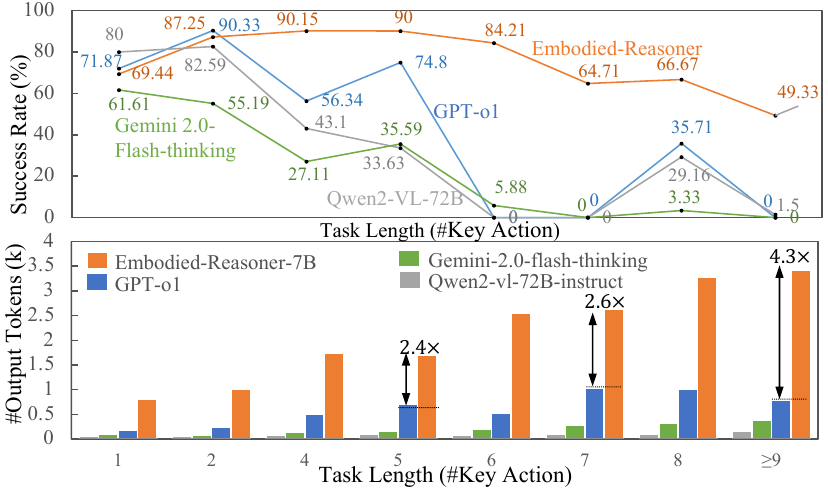}
   \caption{Relations between \emph{task length} and \emph{success rate}, and output \emph{token number}. As task complexity increases, our model generates more reasoning tokens to maintain high success rates.}
   \label{task length}
\end{figure}

\textbf{More robust to long-horizon tasks.} To investigate the impact of the deep-thinking paradigm on embodied search tasks, we count the number of key actions required for each test case. More key actions indicate the task is more complex with more interactions, i.e., long-horizon tasks. As shown in~\cref{task length}, we visualize the relationship between task length (number of key actions), success rate, and the number of output tokens. We observe that as the number of key actions increases, the success rates of baseline models drop significantly—especially when the task exceeds five actions. In contrast, our model remains robust to complex tasks, achieving over 60\% success rate in most scenarios.

\textbf{Spontaneously scale up reasoning tokens for complex embodied tasks.} \cref{task length} (Bottom) shows our \emph{Embodied-Reasoner} utilizes significantly more reasoning tokens for complex searching tasks, nearly five times that of Gemini-2.0-flash-thinking. Besides, as tasks become more complex, the response tokens of our model also grow from 1,000 to 3,500 tokens. We observe that when faced with complex composite searching tasks, \emph{Embodied-Reasoner} engages in significantly longer analysis processes and more deliberate self-reflection. This deep thinking process enables it to plan more efficient search paths and avoid redundant actions, improving the success rate. In contrast, Gemini-2.0-flash-thinking does not show a clear increase in their output tokens, remaining at 1,000 tokens. It suggests it may fail to solve complex embodied tasks by inference time scaling up.

\begin{figure}[t!]
  \centering
   \includegraphics[width=1\linewidth]{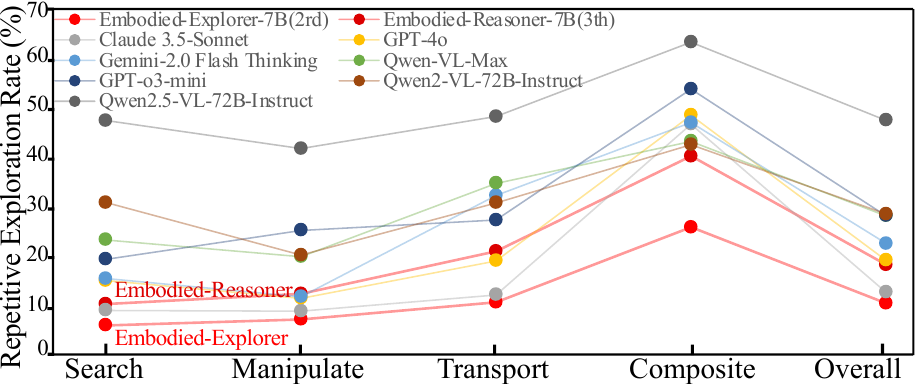}
   \caption{Repetitive Exploration Rate measures repetitive search issues, which are often observed in baselines. Our models reduce repetitive searches by recalling and reflecting on past trajectories.}
   \label{fig_repetitive}
\end{figure}

\textbf{Deep thinking alleviates repetitive searching actions.} We observe that baseline models frequently exhibit repetitive search behaviors. For instance, after inspecting a cabinet, the model may still attempt to check the same cabinet after a few steps. This behavior reflects its weaker temporal reasoning and context awareness abilities in interactive scenarios. To quantify this, we define a \textbf{repetitive exploration rate (RER)}, which measures how often the model navigates to the same area within its trajectory. As shown in \cref{fig_repetitive}, our models (\emph{Embodied-Reasoner / Explorer}) consistently exhibit a significantly lower RER (-50\%) compared to baseline models across all four tasks. For example, in composite tasks, \emph{Embodied-Explorer} achieves the lowest RER of 26\%, while GPT-o3-mini and Qwen2-VL-72B reach 54\% and 43\%, respectively. Compared to \emph{Embodied-Explorer}, \emph{Reasoner} exhibits a slightly higher RER due to its more cautious nature, favoring multiple checks and reflections. In our model's reasoning thoughts, we observe that it frequently recalls past observations, reflects on previously explored actions, and formulates new plans accordingly. These processes enhance its temporal reasoning ability, thereby reducing repetitive search behaviors.


\begin{figure}[t!]
  \centering
   \includegraphics[width=1\linewidth]{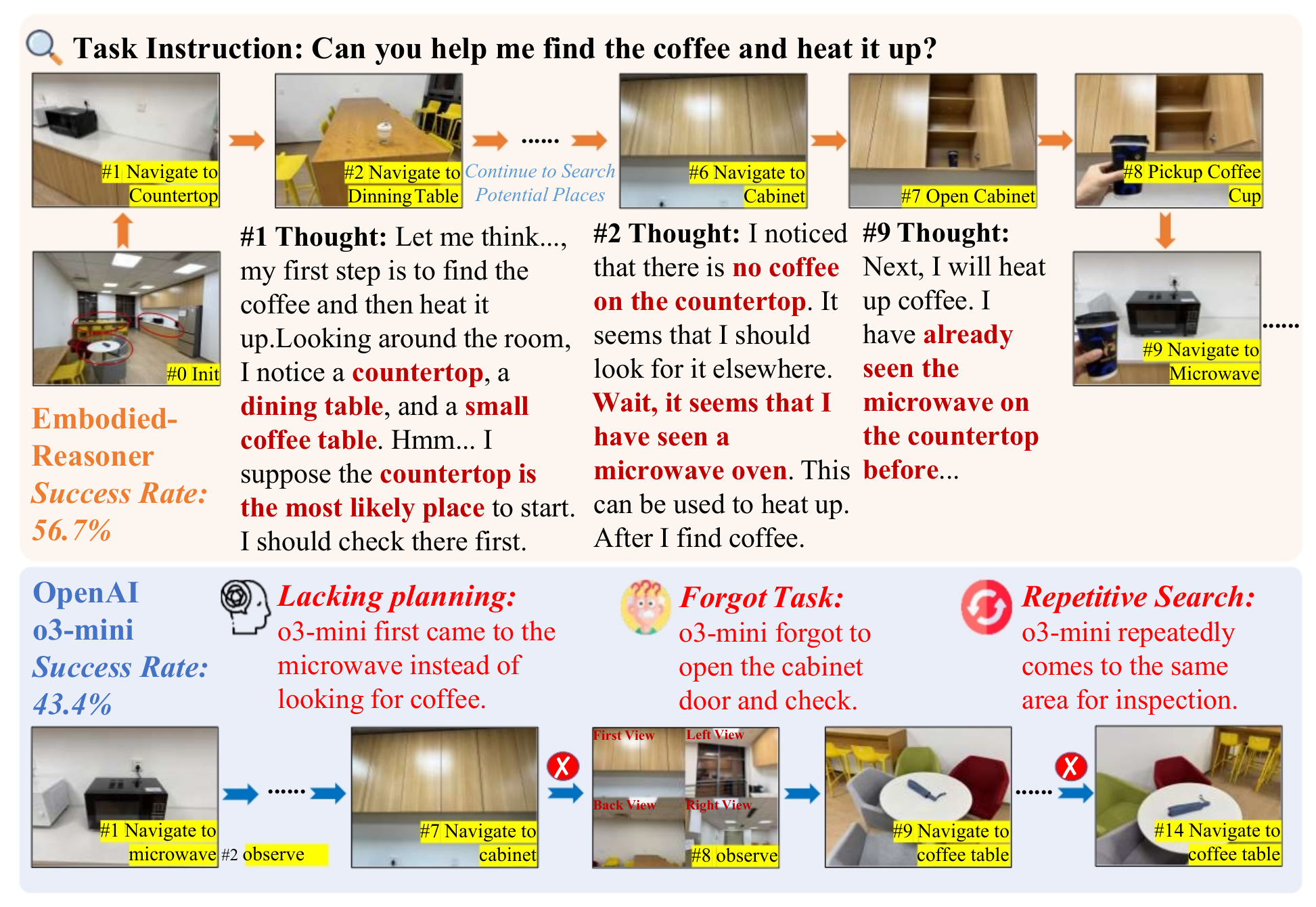}
   \caption{Real-world experiments. Our model achieves a higher success rate (56.7\%) than OpenAI o3-mini (43.4\%) and o1 (50\%).}
   \label{fig_real_world}
\end{figure}

\subsection{Real-world Experiments} 
To evaluate the generalization of our reasoning model, we design a real-world experiment about object searching, covering 30 tasks across three scenes: 6 kitchen tasks, 12 bathroom tasks, and 12 bedroom tasks. During testing, a human operator holds a camera to capture real-time visual input. The model analyzes each image and generates an action command, which the operator executes the actions. 


\cref{fig_real_world} illustrates an example: \emph{“Can you help me find the coffee and heat it up?”} Our model rules out the countertop and dining table after two explorations (steps 1,2), ultimately locating the coffee (\#7) in the cabinet and placing it in the microwave for heating (\#11). However, we observe that OpenAI o3-mini fails to formulate a reasonable plan, heading to the microwave first instead of searching for the coffee. Besides, it frequently forgets to search and exhibits repetitive searching, aligning with our previous analysis. Please refer to~\Cref{real-world experiments result} for detailed results.


\section{Related Works}
\subsection{Large Language Models Reasoning}
Recent o1-style models~\cite{guo2025deepseek,openai-o1,deepmind2025flashthinking,zhang2024o1,zhao2024marco,qwq-32b-preview} have demonstrated powerful reasoning capabilities, significantly enhancing their ability on college-level benchmarks such as mathematics~\cite{rein2023gpqa,hendrycks2021measuring}. Unlike previous efforts to scale up training data and model sizes, these systems involve generating long ``thought'' tokens during inference time, improving the performance of final answers~\cite{guan2025rstar,min2024imitate,zhang2024llama,zhang2024self}. From chain-of-thought (CoT) prompting by humans~\cite{wei2022chain,kojima2022large} to autonomous thinking derived by LLMs themselves~\cite{zhang2024agent}, the reasoning abilities have become increasingly internalized and spontaneous. Besides, the long-thought generation process is quite similar to human thinking activities, with diverse thinking patterns, e.g., step-by-step solving, self-reflection, backtracking, and self-validation~\cite{qin2024o1,huang2025o1}. These characteristics greatly enhance the complex problem-solving capabilities of LLMs. However, most efforts focus on text-based tasks, and the application of deep-thinking paradigms to embodied scenarios remains largely unexplored. This is the focus of our work.

\subsection{Vision-Language Model Reasoning}
Recently, many efforts~\cite{thawakar2025llamav,xu2411llava,xu2025redstar,zhang2024improve,wang2024enhancing}, such as QVQ~\cite{qwenlm2025qvq} and Kimi-1.5~\cite{team2025kimi}, have extended the deep-thinking paradigm to multimodal scenarios by post-training on long-CoT thought or large scale reinforcement learning. However, most visual reasoning models operate in a single-round dialogue setting: processing input images and user's query and generating textual thoughts for a final answer. This limits their applicability in embodied interactive tasks~\cite{ahn2022can,driess2023palm,lin2023text2motion,brohan2023rt,kim2024openvla}, which require handling multi-image or image-text interleaved contexts while generating coherent and logical thoughts across multiple interactions. Besides, embodied scenes differ from mathematical tasks, as they demand long-horizon planning and deliberate reasoning over previous trajectories. Inspired by image-text interleaved understanding and generation~\cite{zhang20252,li2025imagine}, we propose an effective solution to extend general VLMs into embodied reasoning models, including synthesizing interleaved \textit{observation-thought-action} trajectories and bootstrapping VLMs through iterative training.

\subsection{Vision-Language-Action Model}
Some researches integrate language models into embodied tasks~\cite{huang2022language,liang2023code,song2023llm,sarch2023open,ziliotto2024tango}. For instance, agents like PaLM-E\cite{driess2023palm} and other researches\cite{liang2024survey,zhu2023minigpt,ye2024mplug,wu2024embodied,zhao2024see,kak2024embodied,mu2023embodiedgpt,huang2023embodied,qu2025spatialvla,shi2024yell,cai2024rocket,wu2023embodied} leverage the inherent knowledge of large language models (LLMs) and combine robotic data with general visual-language data for training. 
Another part of the research\cite{cui2024survey,chen2024end} emphasizes the integration of language models (LLMs) and visual-anguage models (VLMs) into robotics. Take Chatgpt for robotics\cite{vemprala2024chatgpt} as an example. The study first defines a high - level robotic function library, then creates prompts to guide ChatGPT to achieve its goals.

In addition, some research\cite{wen2023road,wen2024object,zhang2023multimodal,wang2024large} utilizes the capabilities of visual-language models (VLM) to assist traditional strategy training such as reinforcement learning and imitation learning, thus promoting the development of robotic manipulation and navigation\cite{huang2023visual,yokoyama2024hm3d,khanna2024goat,majumdar2024openeqa}. Besides, a large amount of research\cite{zawalski2024robotic,cui2024drive,kannan2024smart,wake2024gpt} has been dedicated to developing generalist robotic strategies, which are trained based on continuously expanding robotic learning datasets. However, most of the existing work directly predicts action sequences without enabling robots to have an autonomous thinking process and this is one of the key focuses of our work.

\section{Conclusions}

In this paper, we propose an embodied reasoning model, \emph{Embodied-Reasoner}, for interactive search tasks that can spontaneously search, reason, and act. To develop this model, we design a data engine that synthesizes 9,390 interactive trajectories in an \emph{Observation-Thought-Action} interleaved format. It encompasses 64K images and 8M thought tokens featuring diverse thinking patterns. We employ a three-stage training approach—imitation learning, rejection sampling tuning, and reflection tuning—to progressively enhance its interaction and reasoning abilities. Extensive evaluations and real-world experiments demonstrate that our model exhibits superior reasoning capabilities.

{
    \small
    \bibliographystyle{ieeenat_fullname}
    \bibliography{main}
}
\appendix
\clearpage
\renewcommand\thefigure{\Alph{section}\arabic{figure}}    
\setcounter{figure}{0}    
\renewcommand\thetable{\Alph{section}\arabic{table}}    
\setcounter{table}{0}  
\clearpage
\setcounter{page}{1}

\section{Experiment Details}\label{experiments details}

\subsection{Real-World Experiments Result}
To evaluate the generalization of our reasoning model, we design a real-world experiment about object searching, covering 30 tasks in three scenes. As shown in ~\cref{real-world experiments result}, \emph{Embodied-Reasoner} demonstrates notable capabilities in real-world environments. In terms of success rate, it outperforms OpenAI o1 by 6.7\%, OpenAI o3-mini by 12.7\%, and Qwen2.5-VL-72B-Instruct by 13.4\%.

\begin{table}[t!]
    \centering 
    \footnotesize  
    \setlength\tabcolsep{4pt} 
    \renewcommand{\arraystretch}{0.9} 
    \begin{tabular}{l|c}
        \toprule[1pt]
        \textbf{Model} & \textbf{Success Rate (\%)}  \\ 
        \toprule[0.5pt]
        Qwen2.5-VL-72B-Instruct & 43.3\\ 
        OpenAI o1 & 50.0\\
        OpenAI o3-mini & 44.0\\  
        Embodied-Reasoner & 56.7\\  
        \bottomrule[1pt]
    \end{tabular}    
    \caption{The results of real-world experiments.}
    \label{real-world experiments result}
\end{table}

\subsection{Repeat Exploration Rate}
The Repeat Exploration Rate (RER)indicates how often the model revisits the same location within its trajectory and is calculated as the number of revisits to previous locations divided by the total number of explorations.

For example, in a task, the model navigated to the following path: $Place_{a}$, $Place_{b}$, $Place_{b}$, $Place_{c}$, $Place_{c}$. The model revisited $Place_{b}$ and $Place_{c}$ once each. Thus, the repeat exploration rate was $40\%$ ($2/5$).




\section{Dataset Details}
\label{sec:dataset-details}




\subsection{Distribution of Task Instructions}
\begin{figure}[t!]
  \centering
   \includegraphics[width=1\linewidth]{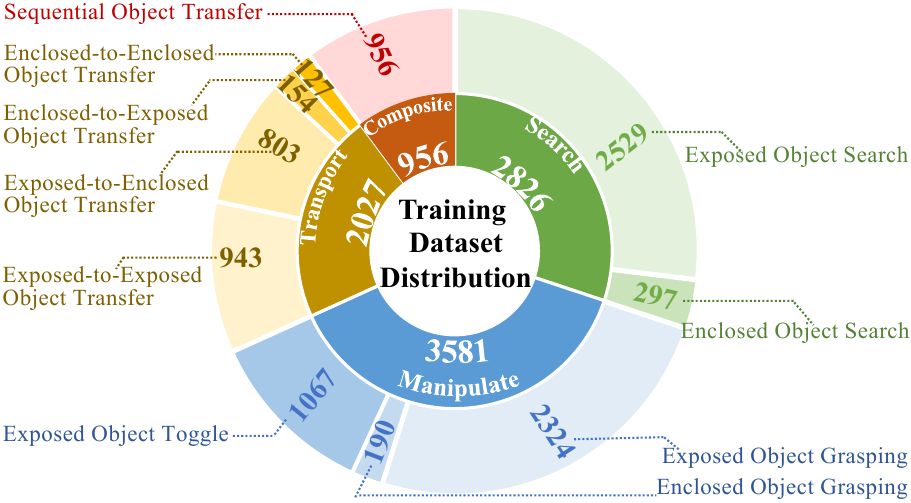}
   \caption{The distribution of the training dataset with \emph{9,390} samples, including \emph{4} task types and \emph{10} sub-task types.}
   \label{trainset distribution}
   
\end{figure}
We synthesize 9,390 unique task instructions along with their \textit{Observation-Thought-Action} trajectories as the training set. The distribution of training tasks is shown in \cref{trainset distribution}, encompassing 4 task types(\emph{Search}, \emph{Manipulate}, \emph{Transport} and \emph{Composite}) and 10 subtask types.

\begin{figure}[t!]
  \centering
   \includegraphics[width=1\linewidth]{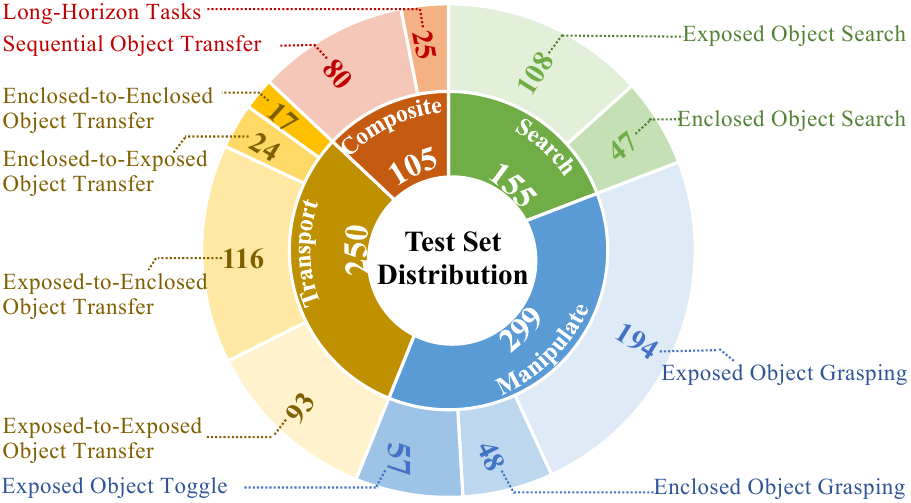}
   \caption{The distribution of the test set with \emph{809} tasks, including \emph{4} task types and \emph{11} sub-task types.}
   \label{testset distribution}
\end{figure}

For evaluation, we curate 809 test cases across 12 novel scenarios distinct from the training environments. The distribution of test tasks is shown in \cref{testset distribution}, covering 4 task types and 11 corresponding subtask types.

\subsection{Distribution of Interaction Types}
\begin{figure}[t!]
  \centering
   \includegraphics[width=1\linewidth]{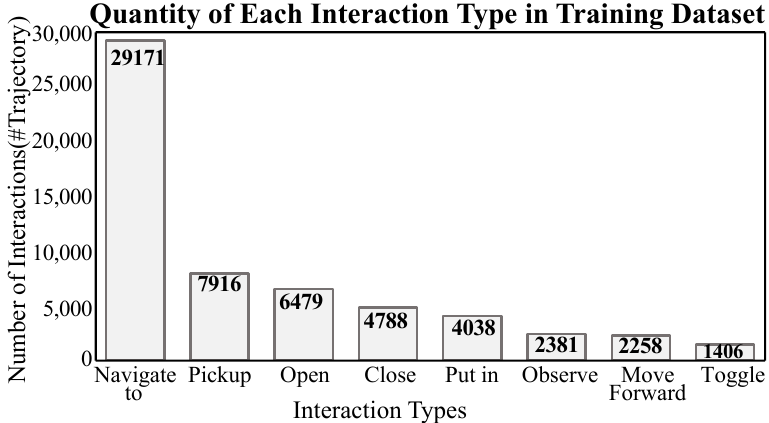}
   \caption{The distribution of the training set interactions, including \emph{8} interaction types in trajectories: \emph{navigate to, pickup, open, close, put in, observe, move forward, and toggle}.}
   \label{trainset interaction distribution}
\end{figure}

In the training set, trajectories consist of eight types of interaction actions: \emph{navigate to, pickup, open, close, put in, observe, move forward, and toggle}. As shown in the \cref{trainset interaction distribution}, the occurrence frequency of each interaction action across all trajectories is illustrated. Among them, the exploration action \emph{navigate to} appears the most frequently, occurring over 29k times.

\begin{figure}[t!]
  \centering
   \includegraphics[width=1\linewidth]{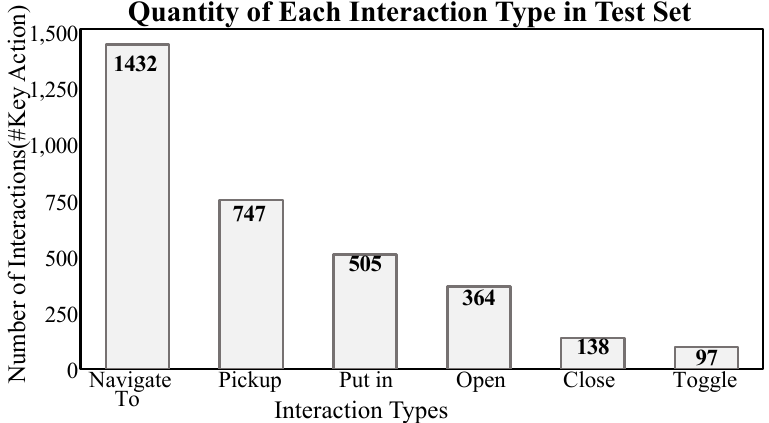}
   \caption{The distribution of the test set interactions, including \emph{6} interaction types in key actions: \emph{navigate to, pickup, open, close, put in, and toggle}.}
   \label{testset interaction distribution}
\end{figure}

In the test set, we manually design instructions and annotate the corresponding key actions and final states. The test tasks involve six types of interactions: \emph{navigate to, pickup, open, close, put in, and toggle}. As seen in the \cref{testset interaction distribution}, \emph{navigate to} also appears significantly more frequently than other key actions.

\subsection{Distribution of Task Length}
\begin{figure}[t!]
  \centering
   \includegraphics[width=1\linewidth]{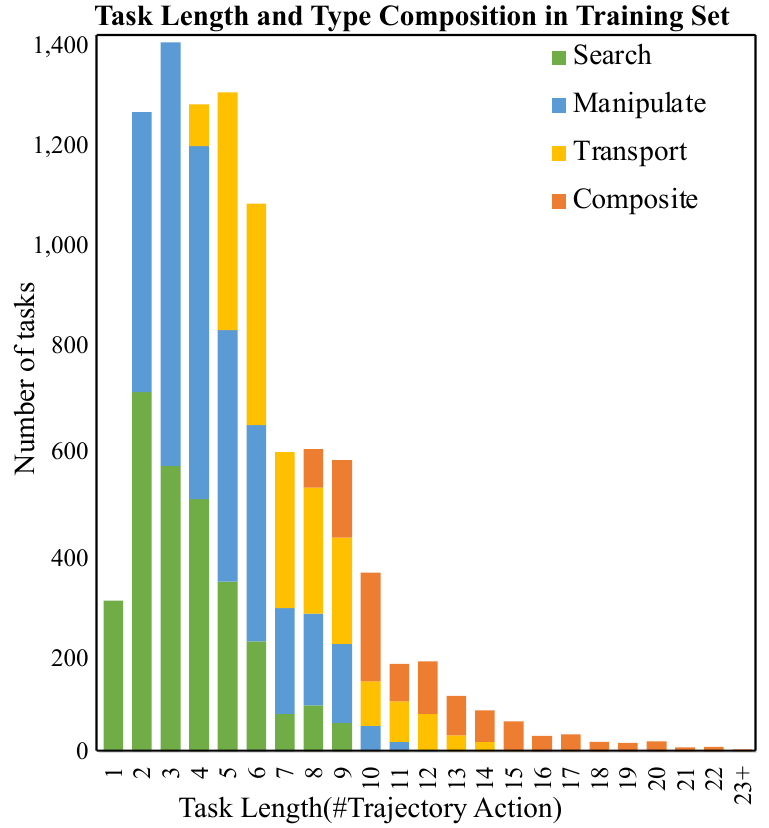}
   \caption{The quantity distribution of trajectory lengths in the training set and the corresponding task type composition, where \emph{Search Task} is mainly within \emph{1-9}, \emph{Manipulate Task} within \emph{2-11}, \emph{Transport Task} within \emph{3-14}, and \emph{Composite Task} above \emph{8}, extending beyond 23.}
   \label{trainset task length}
\end{figure}

In the training set, each trajectory consists of an average of 7.2 interactions with the environment (e.g., \emph{navigate}, \emph{pickup}). For the four task types: \emph{Search}, \emph{Manipulate}, \emph{Transport}, and \emph{Composite}. Due to varying task complexity, the corresponding trajectory lengths also differ. As shown in the \cref{trainset task length}, \emph{Search} tasks tend to have shorter trajectories, typically ranging from 1 to 9, while the more complex \emph{Composite} tasks generate the longest trajectories, usually exceeding 8 and reaching beyond 23.

\begin{figure}[t!]
  \centering
   \includegraphics[width=1\linewidth]{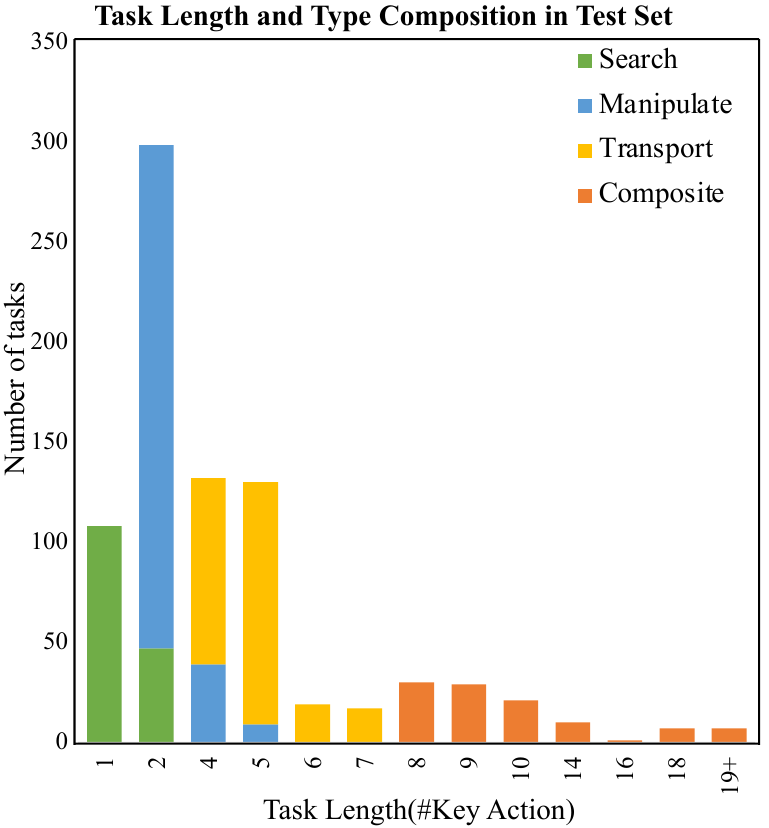}
   \caption{The quantity distribution of key action lengths in the test set and the corresponding task type composition, where \emph{Search Task} is mainly within 1-2, \emph{Manipulate Task} within 2, 4-5, \emph{Transport Task} within 4-7, and \emph{Composite Task} above 8, extending beyond 19.}
   \label{testset task length}
\end{figure}

Similarly, in the test set, as shown in the \cref{testset task length}, the more complex \emph{Composite} tasks also exhibit the longest key action sequences, usually exceeding 8 and reaching beyond 19.

\subsection{Distribution of Object Categories}

\begin{figure}[t!]
  \centering
   \includegraphics[width=1\linewidth]{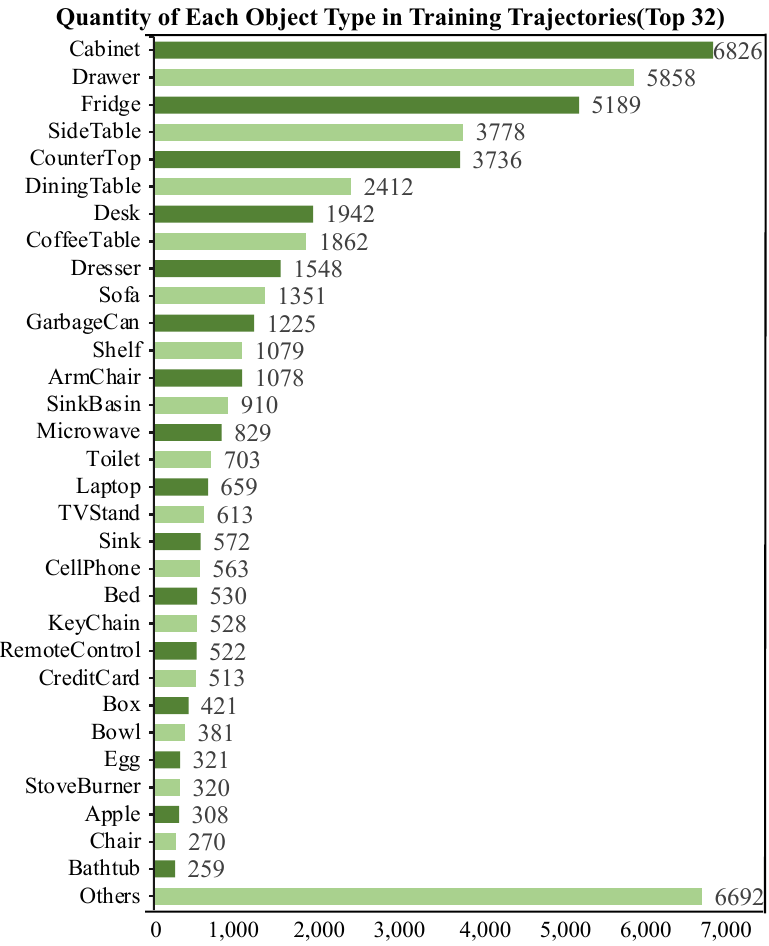}
   \caption{The quantity distribution of the top 32 object types in the training dataset's trajectories, with \emph{Others} representing the remaining 62 categories, such as \emph{Bread, Book, DeskLamp}, etc.}
   \label{trainset object count}
\end{figure}
\begin{figure}[t!]
  \centering
   \includegraphics[width=1\linewidth]{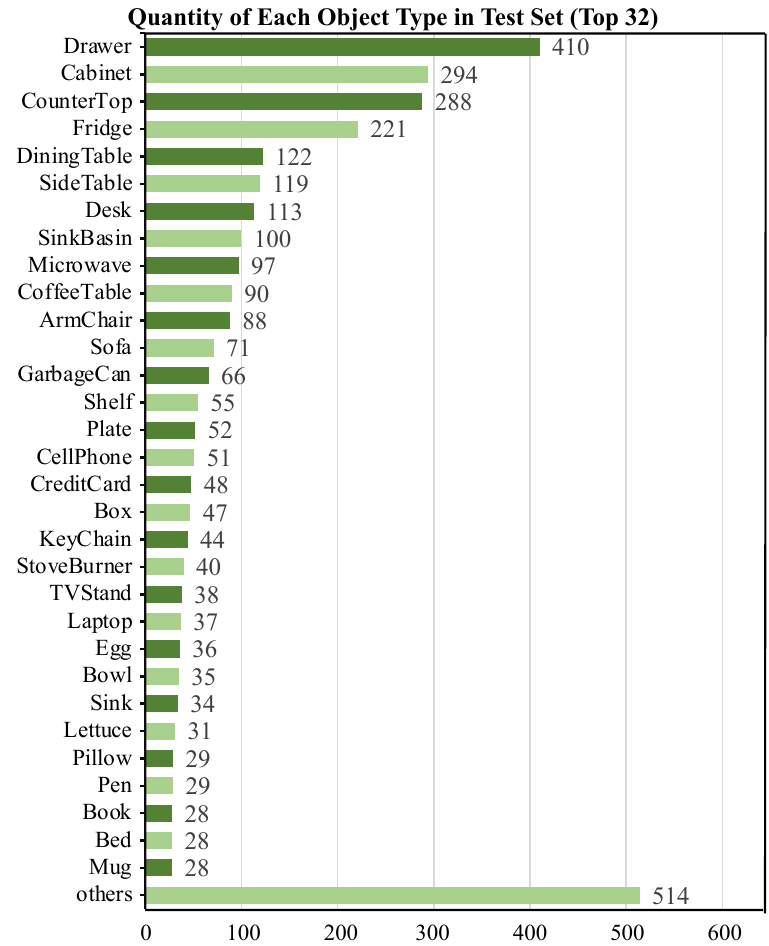}
   \caption{The quantity distribution of the top 32 object types in the test set's key actions, with \emph{Others} representing the remaining 44 categories, such as \emph{Watch, Pencil, Cup}, etc.}
   \label{testset object count}
\end{figure}

Our training dataset includes 107 indoor scenes with diverse functions and layouts (kitchens, living rooms, bedrooms, and bathrooms), featuring over 2,100 interactive objects (e.g., eggs, laptops) and 2,600 containers (e.g., refrigerators, drawers). Across the 9,390 unique task instructions, trajectories involve interactions with these objects and containers, with the top 32 most frequently explored and interacted object categories shown in \cref{trainset object count}. In the 12 distinct test set scenes, key actions correspond to different objects, with the top 32 most frequently involved containers and interactive objects illustrated in \cref{testset object count}.

\subsection{Description of Sub-task Types}
Our four daily tasks: \textit{Search}, \textit{Manipulate}, \textit{Transport}, and \textit{Composite} can be further divided into corresponding sub-tasks based on the types of objects involved. Specifically, the \textit{Search Task} can be categorized into Exposed Object Search and Enclosed Object Search. The \textit{Manipulate Task} can be divided into Exposed Object Grasping, Enclosed Object Grasping, and Exposed Object Toggle. The \textit{Transport Task} can be classified into Exposed-to-Exposed Object Transfer, Exposed-to-Enclosed Object Transfer, Enclosed-to-Exposed Object Transfer, and Enclosed-to-Enclosed Object Transfer. Finally, the \textit{Composite Task} can be divided into Sequential Object Transfer and Long-Term Complex Task.

\textbf{Exposed Object Search.} This task is defined as searching for items within a room environment. The target items are located on the surface and there is no need to open any containers. For example, if there is an apple placed on the table in the room, the corresponding task description is "Please find the apple in the room".

\textbf{Enclosed Object Search.} This type of task refers to searching for items in a room where the target items are inside containers, and the containers need to be opened during the task execution. For instance, if there is an egg in the refrigerator in the room, then the task description is "Please find the egg in the room".

\textbf{Exposed Object Grasping.} This type of task requires obtaining a specific item in a room, and the target item is on the surface without the need to open any containers. For example, when there is a cup on the table in the room, the corresponding task description is "Please pick up the cup in the room".

\textbf{Enclosed Object Grasping.} This task is to obtain an item located inside a container in a room, and the container needs to be opened during the execution. For example, if there is a bowl in the cabinet in the room, the corresponding task description is "Please pick up the bowl in the room".

\textbf{Exposed Object Toggle.} This task is to perform switch operations on items located on the surface in a room, without the need to open any containers. For example, if there is a coffee machine on the table in the room, the corresponding task description is "Please start the coffee machine".

\textbf{Exposed-to-Exposed Object Transfer.} his task requires moving an item located on the surface (without opening any containers) to another position in the room. For example, if there is a spoon on the table in the room, the task description is "Please pick up the spoon in the room and place it in the sink".

\textbf{Exposed-to-Enclosed Object Transfer.} This task is defined as moving an item located on the surface (without opening any containers) into a closed container in the room. The container needs to be opened to complete the placement. For example, if there is a credit card on the table in the room, the task description is "Please pick up the credit card in the room and place it in the drawer".

\textbf{Enclosed-to-Exposed Object Transfer.} This task refers to moving an item located inside a container to another position after opening the container, finding and picking up the item. For example, if there is a dishwashing sponge in the cabinet, the corresponding task description is "Please pick up the dishwashing sponge in the room and place it on the table".

\textbf{Enclosed-to-Enclosed Object Transfer.} This task is defined as moving an item located inside one container to another closed container in the room. Two containers need to be opened during the execution. For example, if there is a loaf of bread in the refrigerator in the room, the task description is "Please pick up the bread in the room and place it in the cabinet".

\textbf{Sequential Object Transfer.} This task requires moving different items to specified locations in a room in a specific order. It is composed of two of the three task types: Exposed-to-Exposed Object Transfer, Exposed-to-Enclosed Object Transfer, and Enclosed-to-Exposed Object Transfer. For example, if there are apples on the table and eggs in the cabinet in the room, the task description is "First, find the apples in the room and place them in the sink (involving Exposed-to-Exposed Object Transfer), then find the eggs and place them on the table (involving Enclosed-to-Exposed Object Transfer)".

\textbf{Long Term Complex Task.} This task involves performing a series of ordered and complex long-range operations in a room. It is composed of four of the five types: Exposed Object Toggle, Exposed-to-Exposed Object Transfer, Enclosed-to-Exposed Object Transfer, Enclosed-to-Enclosed Object Transfer, and Exposed-to-Enclosed Object Transfer. For example, if there are potatoes in the refrigerator and a mug on the table in the room, the task description is: First, please find the potatoes in the room and wash them (this process involves Exposed-to-Enclosed Object Transfer and Enclosed Object Transfer), then put the potatoes in the microwave to heat (involving Exposed-to-Enclosed Object Transfer and Exposed Object Toggle), then find the mug and place it on the coffee machine to get coffee and then place the mug with coffee on the table (involving Exposed Object Transfer and Exposed Object Toggle), and finally place the potatoes from the microwave on the table (involving Enclosed Object Transfer).

\section{Detailed Task Templates and Constraints}
We design multiple task templates for each task. It ensures synthesized instruction's validity. Templates and constraints are shown in \cref{tab:detailed task templates}.

\begin{table*}[h]\small
  \centering
  \label{tab:detailed task templates}
  \begin{tabular}{l|l|l|l|l|l}
    \toprule[1pt]
    \makecell[c]{Task Types} & 
    \makecell[c]{Sub-Task Types} & 
    \makecell[c]{Templates} & 
    \makecell[c]{Constraint Check} & 
    \makecell[c]{Case} & 
    \makecell[c]{Affiliation and Attribute} \\
    
    \cline{1-6}
    \multirow{2}{*}{Search} 
    & \makecell[l]{Exposed Object \\ Search} 
    & find \textcolor{cyan}{A} 
    & \makecell[l]{Pickupable(\textcolor{cyan}{A}) \\ $\land$ \\ $\neg$ Openable(\textcolor{brown}{Parent(A)})} 
    & \makecell[l]{
    \textbf{Task:} Could you please \\find the \textcolor{cyan}{Apple} in the \\room? \\ 
    \textbf{Key Action Sequences:} \\ 
    \textcolor{blue}{navigate to} \textcolor{brown}{CounterTop} \\ 
    \textcolor{blue}{end}} 
    & \makecell[l]{\includegraphics[width=4cm,height=1.7cm]{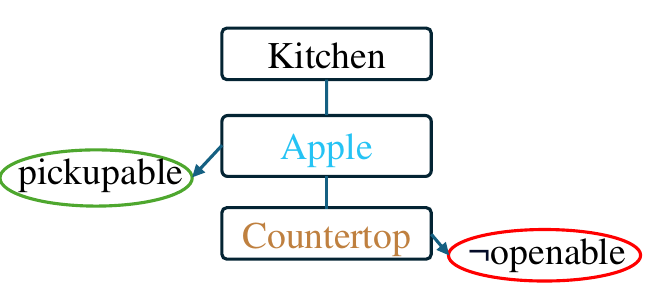}} 
    \\
    \cline{2-6}
    & \makecell[l]{Enclosed Object\\ Search} 
    & find \textcolor{cyan}{A} 
    & \makecell[l]{Pickupable(\textcolor{cyan}{A}) \\ $\land$ \\ Openable(\textcolor{brown}{Parent(A)})} 
    & \makecell[l]{
    \textbf{Task:} Could you please \\find the \textcolor{cyan}{Apple} in the \\room? \\ \textbf{Key Action Sequences:} \\ \textcolor{blue}{navigate to} \textcolor{brown}{Fridge} \\ \textcolor{blue}{open} \textcolor{brown}{Fridge} \\ \textcolor{blue}{end}} 
    & \makecell[l]{\includegraphics[width=4cm,height=1.7cm]{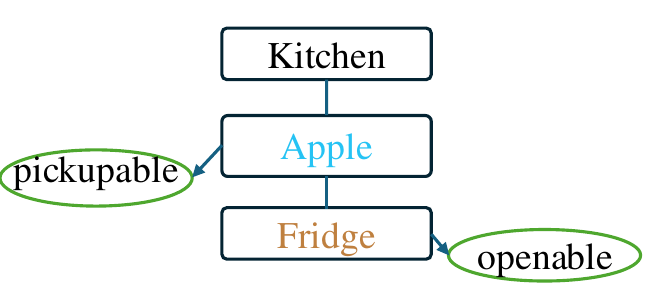}}
    \\

    \cline{1-6}
    \multirow{3}{*}{Manipulate} 
    & \makecell[l]{Exposed Object\\ Toggle}
    & toggle \textcolor{cyan}{A}
    & \makecell[l]{Toggleable(\textcolor{cyan}{A}) \\ $\land$ \\ $\neg$ Openable(\textcolor{brown}{Parent(A)})}
    & \makecell[l]{
    \textbf{Task:} Would you mind \\powering on the \textcolor{cyan}{Laptop} \\for me? \\ \textbf{Key Action Sequences:} \\ 
    \textcolor{blue}{navigate to} \textcolor{brown}{Desk} \\ 
    \textcolor{blue}{toggle} \textcolor{cyan}{Laptop},
    \textcolor{blue}{end}}
    & \makecell[l]{\includegraphics[width=4cm,height=1.7cm]{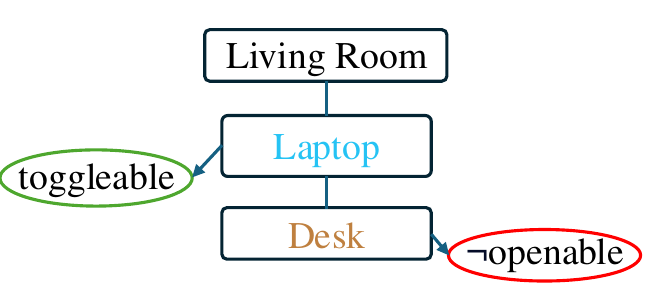}} 
    \\
    \cline{2-6}
    & \makecell[l]{Exposed Object \\Grasping} 
    & pickup \textcolor{cyan}{A}
    & \makecell[l]{Pickupable(\textcolor{cyan}{A}) \\ $\land$ \\ $\neg$ Openable(\textcolor{brown}{Parent(A)})} 
    & \makecell[l]{
    \textbf{Task:} I want to pick up \\a \textcolor{cyan}{CreditCard} from the \\room, can you help me? \\ 
    \textbf{Key Action Sequences:} \\ 
    \textcolor{blue}{navigate to} \textcolor{brown}{SideTable} \\ 
    \textcolor{blue}{pickup} \textcolor{cyan}{CreditCard} \\ 
    \textcolor{blue}{end}} 
    & \makecell[l]{\includegraphics[width=4cm,height=1.7cm]{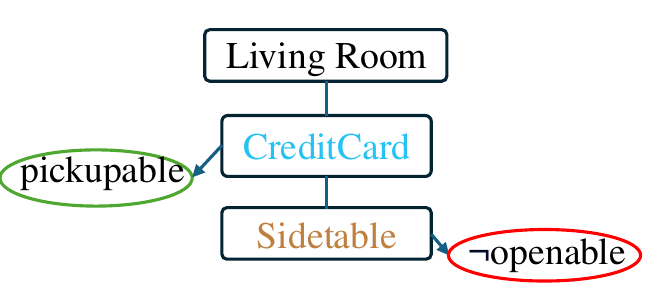}}
    \\
    \cline{2-6}
    & \makecell[l]{Enclosed Object\\ Grasping} 
    & pickup \textcolor{cyan}{A} & \makecell[l]{Pickupable(\textcolor{cyan}{A}) \\ $\land$ \\ Openable(\textcolor{brown}{Parent(A)})} 
    & \makecell[l]{
    \textbf{Task:} Would it be \\possible for you to\\ pick up a \textcolor{cyan}{CreditCard}\\ from the room? \\ 
    \textbf{Key Action Sequences:} \\ 
    \textcolor{blue}{navigate to} \textcolor{brown}{Drawer} \\
    \textcolor{blue}{open} \textcolor{brown}{Drawer} \\
    \textcolor{blue}{pickup} \textcolor{cyan}{CreditCard},
    \textcolor{blue}{end}} 
    & \makecell[l]{\includegraphics[width=4cm,height=1.7cm]{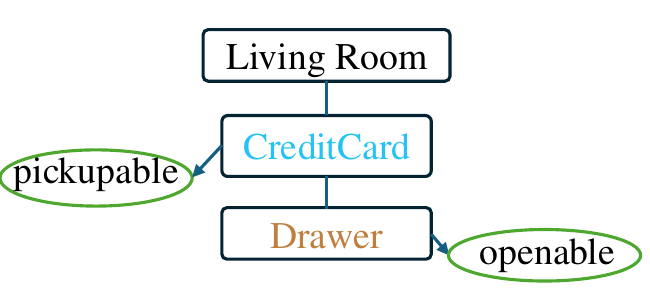}}
    \\

    \cline{1-6}
    \multirow{2}{*}{Transport} 
    & \makecell[l]{Exposed-to-\\Exposed \\Object Transfer} 
    & \makecell[l]{pickup \textcolor{cyan}{A} \\ put in \textcolor{brown}{B}}
    & \makecell[l]{Pickupable(\textcolor{cyan}{A}) \\ $\land$ \\ $\neg$ Openable(\textcolor{brown}{Parent(A)})\\ $\land$ \\ $\neg$ Openable(\textcolor{brown}{B})} 
    & \makecell[l]{
    \textbf{Task:} Could you please\\ put the \textcolor{cyan}{AlarmClock} on \\the \textcolor{brown}{Shelf}? \\ 
    \textbf{Key Action Sequences:} \\ 
    \textcolor{blue}{navigate to} \textcolor{brown}{Sidetable} \\ 
    \textcolor{blue}{pickup} \textcolor{cyan}{AlarmClock} \\ 
    \textcolor{blue}{navigate to} \textcolor{brown}{Shelf} \\ 
    \textcolor{blue}{put in} \textcolor{brown}{Shelf},
    \textcolor{blue}{end}} 
    & \makecell[l]{\includegraphics[width=4cm,height=2cm]{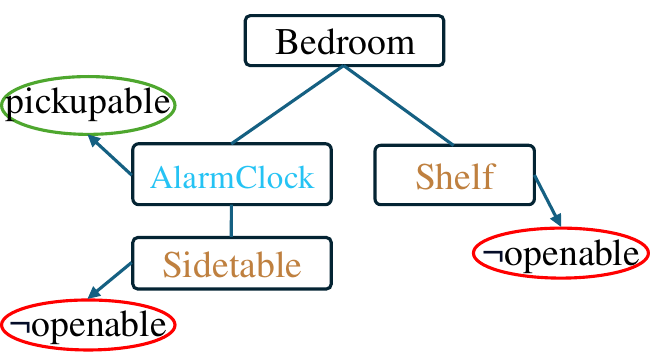}} 
    \\
    \cline{2-6}
    & \makecell[l]{Exposed-to-\\Enclosed \\Object Transfer} 
    & \makecell[l]{pickup \textcolor{cyan}{A} \\ put in \textcolor{brown}{B}}
     & \makecell[l]{Pickupable(\textcolor{cyan}{A}) \\ $\land$ \\ $\neg$ Openable(\textcolor{brown}{Parent(A)})\\ $\land$ \\ Openable(\textcolor{brown}{B})} 
    & \makecell[l]{
    \textbf{Task:} Would you mind \\placing the \textcolor{cyan}{Bowl} in \\the \textcolor{brown}{Cabinet}, please? \\ 
    \textbf{Key Action Sequences:} \\ 
    \textcolor{blue}{navigate to} \textcolor{brown}{CounterTop} \\ 
    \textcolor{blue}{pickup} \textcolor{cyan}{Bowl} \\ 
    \textcolor{blue}{navigate to} \textcolor{brown}{Cabinet} \\
    \textcolor{blue}{open} \textcolor{brown}{Cabinet} \\ 
    \textcolor{blue}{put in} \textcolor{brown}{Cabinet},
    \textcolor{blue}{end}} 
    & \makecell[l]{\includegraphics[width=4cm,height=2cm]{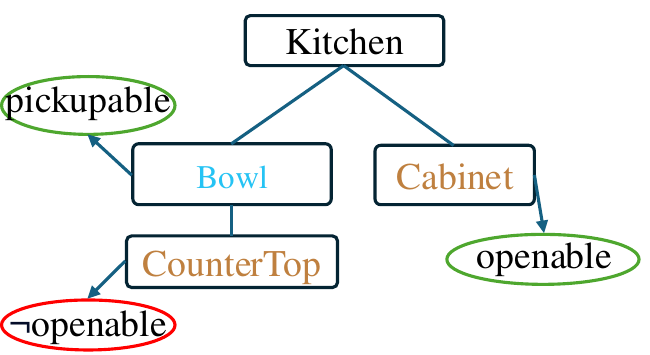}}
    \\
    \toprule[1pt]
    
  \end{tabular}
\end{table*}

\begin{table*}[t]\small
  \centering
  \begin{tabular}{l|l|l|l|l|l}
    \toprule[1pt]
    \makecell[c]{Task Types} & 
    \makecell[c]{Sub-Task Types} & 
    \makecell[c]{Templates} & 
    \makecell[c]{Constraint Check} & 
    \makecell[c]{Case} & 
    \makecell[c]{Affiliation and Attribute} \\

    \cline{1-6}
    \multirow{2}{*}{Transport} 
    & \makecell[l]{Enclosed-to-\\Exposed \\Object Transfer} 
    & \makecell[l]{pickup \textcolor{cyan}{A} \\ put in \textcolor{brown}{B}}
    & \makecell[l]{Pickupable(\textcolor{cyan}{A}) \\ $\land$ \\Openable(\textcolor{brown}{Parent(A)})\\ $\land$ \\ $\neg$ Openable(\textcolor{brown}{B})} 
    & \makecell[l]{
    \textbf{Task:} Is it okay to put \\the \textcolor{cyan}{Candle} on the \\\textcolor{brown}{Bathtub}? \\ 
    \textbf{Key Action Sequences:} \\ 
    \textcolor{blue}{navigate to} \textcolor{brown}{Cabinet} \\ 
    \textcolor{blue}{open} \textcolor{brown}{Cabinet} \\ 
    \textcolor{blue}{pickup} \textcolor{cyan}{Candle} \\ 
    \textcolor{blue}{close} \textcolor{brown}{Cabinet} \\
    \textcolor{blue}{navigate to} \textcolor{brown}{Bathtub} \\ 
    \textcolor{blue}{put in} \textcolor{brown}{Bathtub} \\
    \textcolor{blue}{end}} 
    & \makecell[l]{\includegraphics[width=4cm,height=2cm]{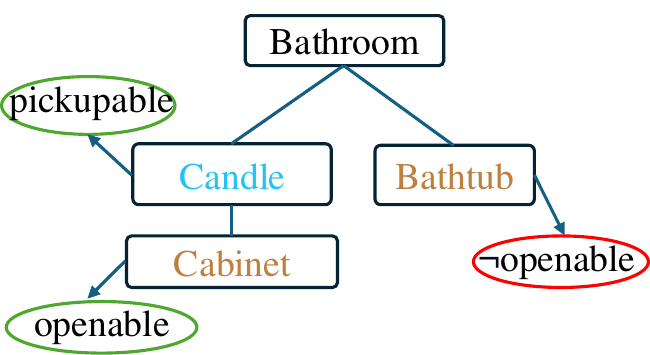}} 
    \\
    \cline{2-6}
    & \makecell[l]{Enclosed-to-\\Enclosed \\Object Transfer} 
    & \makecell[l]{pickup \textcolor{cyan}{A} \\ put in \textcolor{brown}{B}}
     & \makecell[l]{Pickupable(\textcolor{cyan}{A}) \\ $\land$ \\Openable(\textcolor{brown}{Parent(A)})\\ $\land$ \\ Openable(\textcolor{brown}{B})} 
    & \makecell[l]{
    \textbf{Task:} May I ask you to \\put the \textcolor{cyan}{Potato} in the\\ \textcolor{brown}{Microwave}? \\ 
    \textbf{Key Action Sequences:} \\ 
    \textcolor{blue}{navigate to} \textcolor{brown}{Fridge} \\
    \textcolor{blue}{open} \textcolor{brown}{Fridge} \\
    \textcolor{blue}{pickup} \textcolor{cyan}{Potato} \\ 
    \textcolor{blue}{close} \textcolor{brown}{Fridge} \\
    \textcolor{blue}{navigate to} \textcolor{brown}{Microwave} \\
    \textcolor{blue}{open} \textcolor{brown}{Microwave} \\ 
    \textcolor{blue}{put in} \textcolor{brown}{Microwave} \\
    \textcolor{blue}{end}} 
    & \makecell[l]{\includegraphics[width=4cm,height=2cm]{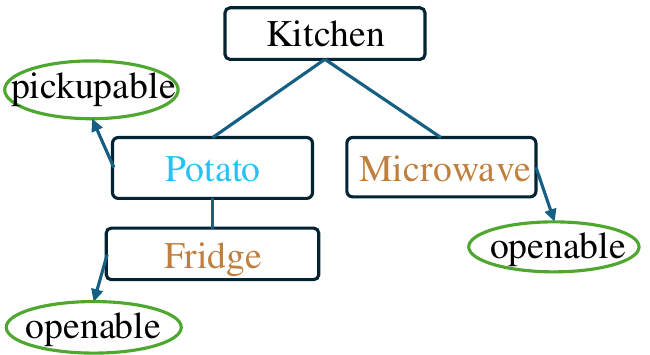}}
    \\

    \cline{1-6}
    \multirow{2}{*}{\makecell[l]{Composite \\task}}
    & \makecell[l]{Sequential \\Object Transfer} 
    & \makecell[l]{first, \\pickup \textcolor{cyan}{A1} \\ put in \textcolor{brown}{B1} \\ then, \\pickup \textcolor{cyan}{A2}\\ put in \textcolor{brown}{B2}}
     & \makecell[l]{Pickupable(\textcolor{cyan}{A1}) \\ $\land$ \\ $\neg$ Openable(\textcolor{brown}{Parent(A1)})\\ $\land$ \\ Openable(\textcolor{brown}{B1})\\ $\land$ \\Pickupable(\textcolor{cyan}{A2}) \\ $\land$ \\ $\neg$ Openable(\textcolor{brown}{Parent(A2)})\\ $\land$ \\ Openable(\textcolor{brown}{B2})\\$\land$ \\Different(\textcolor{cyan}{A1}, \textcolor{cyan}{A2})} 
    & \makecell[l]{
    \textbf{Task:} Could you please\\first place the\\ \textcolor{cyan}{TeddyBear} on the \\\textcolor{brown}{CoffeeTable}, and then\\ place the \textcolor{cyan}{Pen} on\\ the \textcolor{brown}{GarbageCan}? \\ 
    \textbf{Key Action Sequences:} \\ 
    \textcolor{blue}{navigate to} \textcolor{brown}{Bed} \\ 
    \textcolor{blue}{pickup} \textcolor{cyan}{TeddyBear} \\ 
    \textcolor{blue}{navigate to} \textcolor{brown}{CoffeeTable} \\
    \textcolor{blue}{put in} \textcolor{brown}{CoffeeTable} \\
    \textcolor{blue}{navigate to} \textcolor{brown}{Desk} \\ 
    \textcolor{blue}{pickup} \textcolor{cyan}{Pen} \\ 
    \textcolor{blue}{navigate to} \textcolor{brown}{GarbageCan} \\
    \textcolor{blue}{put in} \textcolor{brown}{GarbageCan} \\
    \textcolor{blue}{end}} 
    & \makecell[l]{\includegraphics[width=4cm,height=4cm]{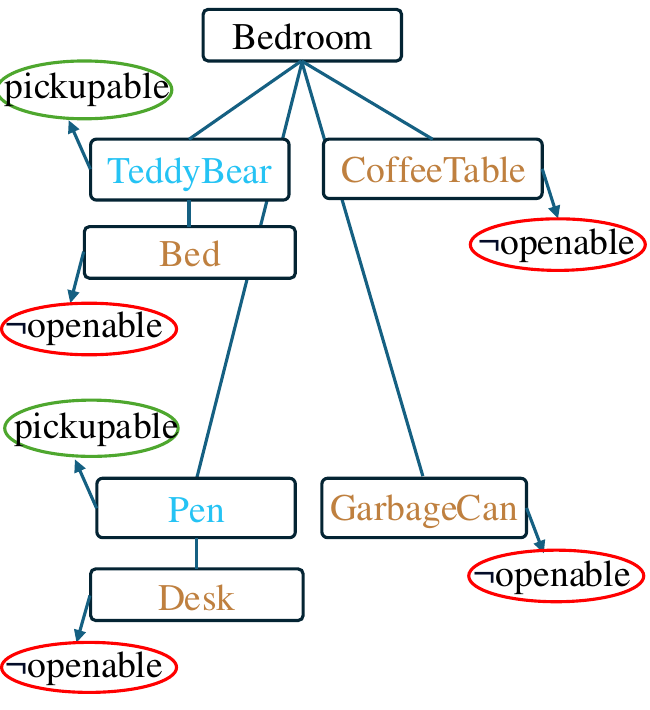}}
    \\
    \cline{2-6}
    & \makecell[l]{Long-Term \\Complex Task} 
    & \makecell[c]{------}
     & \makecell[c]{------} 
    & \multicolumn{2}{l}{\makecell[l]{
    \textbf{Task:} Could you put the \textcolor{cyan}{bread} in the \textcolor{brown}{refrigerator} and then \\take the \textcolor{cyan}{apple} out of the \textcolor{brown}{refrigerator}, wash it, and place it\\ on a \textcolor{brown}{plate}?\\ 
    \textbf{Key Action Sequences:} \\ 
    \textcolor{blue}{navigate to} \textcolor{brown}{CounterTop}, 
    \textcolor{blue}{pickup} \textcolor{cyan}{Bread}, \\
    \textcolor{blue}{navigate to} \textcolor{brown}{Fridge}, 
    \textcolor{blue}{open} \textcolor{brown}{Fridge}\\
    \textcolor{blue}{put in} \textcolor{brown}{Fridge},
    \textcolor{blue}{pickup} \textcolor{cyan}{Apple} \\
    \textcolor{blue}{navigate to} \textcolor{brown}{SinkBasin}, 
    \textcolor{blue}{put in} \textcolor{brown}{SinkBasin}\\
    \textcolor{blue}{toggle} \textcolor{brown}{Facuet},
     \textcolor{blue}{pickup} \textcolor{cyan}{Apple}\\
    \textcolor{blue}{navigate to} \textcolor{brown}{Cabinet},
    \textcolor{blue}{open} \textcolor{brown}{Cabinet}\\
    \textcolor{blue}{put} \textcolor{brown}{Plate},
    \textcolor{blue}{close} \textcolor{brown}{Cabinet}\\
    \textcolor{blue}{end}}}\\
    \toprule[1pt]
    
  \end{tabular}
\end{table*}

\section{Detailed Prompts}
We provide detailed interaction prompt designs for our evaluation framework in \cref{{evaluate framework prompt}}.
\begin{figure*} 
\label{evaluate framework prompt}
\lstset{
    framesep=20pt,
    rulesep=10pt,
    backgroundcolor=\color[RGB]{245,245,244},
    breaklines=true,
    breakindent=0pt,
    basicstyle=\ttfamily\small,
    escapeinside={(*@}{@*)}, 
}
\begin{lstlisting}
(*@\textbf{System prompt: }@*) 
You are a robot in given room. You need to complete the tasks according to human instructions. We provide an Available_Actions set and the corresponding explanations for each action. Each step, you should select one action from Available_Actions.

(*@\textbf{Initialization prompt: }@*)
<image>This is an image from your frontal perspective. Please select an action from the Available_Actions and fill in the arguments.
Task: {(*@\color{blue}{taskname}@*)}
Available_Actions: {{
"navigate to <object>": Move to the object.
"pickup <object>": Pick up the object.
"put in <object>": Put the item in your hand into or on the object.
"toggle <object>": Switch the object on or off.
"open <object>": Open the object (container), and you will see inside the object.
"close <object>": Close the object.
"observe": You can obtain image of your directly rear, left, and right perspectives.
"move forward": Move forward to see more clearly.
"end": If you think you have completed the task, please output "end".}}
Before making each decision, you can think, plan, and even reflect step by step, and then output your final action.
Your final action must strictly follow format: <DecisionMaking>Your Action</DecisionMaking>, for example, <DecisionMaking>observe</DecisionMaking>.

(*@\textbf{Interaction prompt: }@*)
After executing your previous {(*@\color{blue}{action}@*)} , you get this new image above.
To complete your task, you can think step by step at first and then output your new action from the Available_Actions.
Your action must strictly follow format: <DecisionMaking>Your Action</DecisionMaking>, for example, <DecisionMaking>observe</DecisionMaking>.

(*@\textbf{Interaction feedback prompt 1: }@*)
<|feedback|>Action: {(*@\color{blue}{action}@*)} is illegal, {(*@\color{blue}{object}@*)} is the most relevant item in this room and {(*@\color{blue}{action}@*)}. Object: {(*@\color{blue}{object}@*)} is not currently navigable, you can try "navigate to <object>" to reach nearby, larger objects for closer observation.

(*@\textbf{Interaction feedback prompt 2: }@*)
<|feedback|>Action: {(*@\color{blue}{object}@*)} is illegal, Object: {(*@\color{blue}{object}@*)} is currently unavailable for interaction. Possible situations include: {(*@\color{blue}{object}@*)} does not exist in your current view; you are too far away from {(*@\color{blue}{object}@*)}; the {(*@\color{blue}{object}@*)} cannot perform operation {(*@\color{blue}{action}@*)}.\nYou can try "move forward" to approach the target object or "navigate to <object>" to reach nearby, larger objects for closer inspection.

(*@\textbf{Interaction feedback prompt 3: }@*)
<|feedback|>Action: {(*@\color{blue}{action}@*)} is illegal, the name of the navigated object doesn't quite match the obejct in the image, please try navigating to another object first.

(*@\textbf{Interaction feedback prompt 4: }@*)
<|feedback|>Action: {(*@\color{blue}{action}@*)} is illegal, the name of the object doesn't quite match the obejct in the image, Please try interacting with another object or navigating to another object.
\end{lstlisting}
\end{figure*}

\section{Case Study}
As shown in~\Cref{case_our1}, ~\Cref{o1}, and ~\Cref{case_our_real_world}, we provide three cases. The first two illustrate different trajectory performances of Embodied-Reasoner and GPT-o1 for the same task, while the third presents Embodied-Reasoner’s performance in a real-world scenario.

\subsection{Embodied-Reasoner's Action Trajectories}\label{Embodied-Reasoner}
\begin{figure*}[t!]
  \centering
  \includegraphics[width=0.9\linewidth]{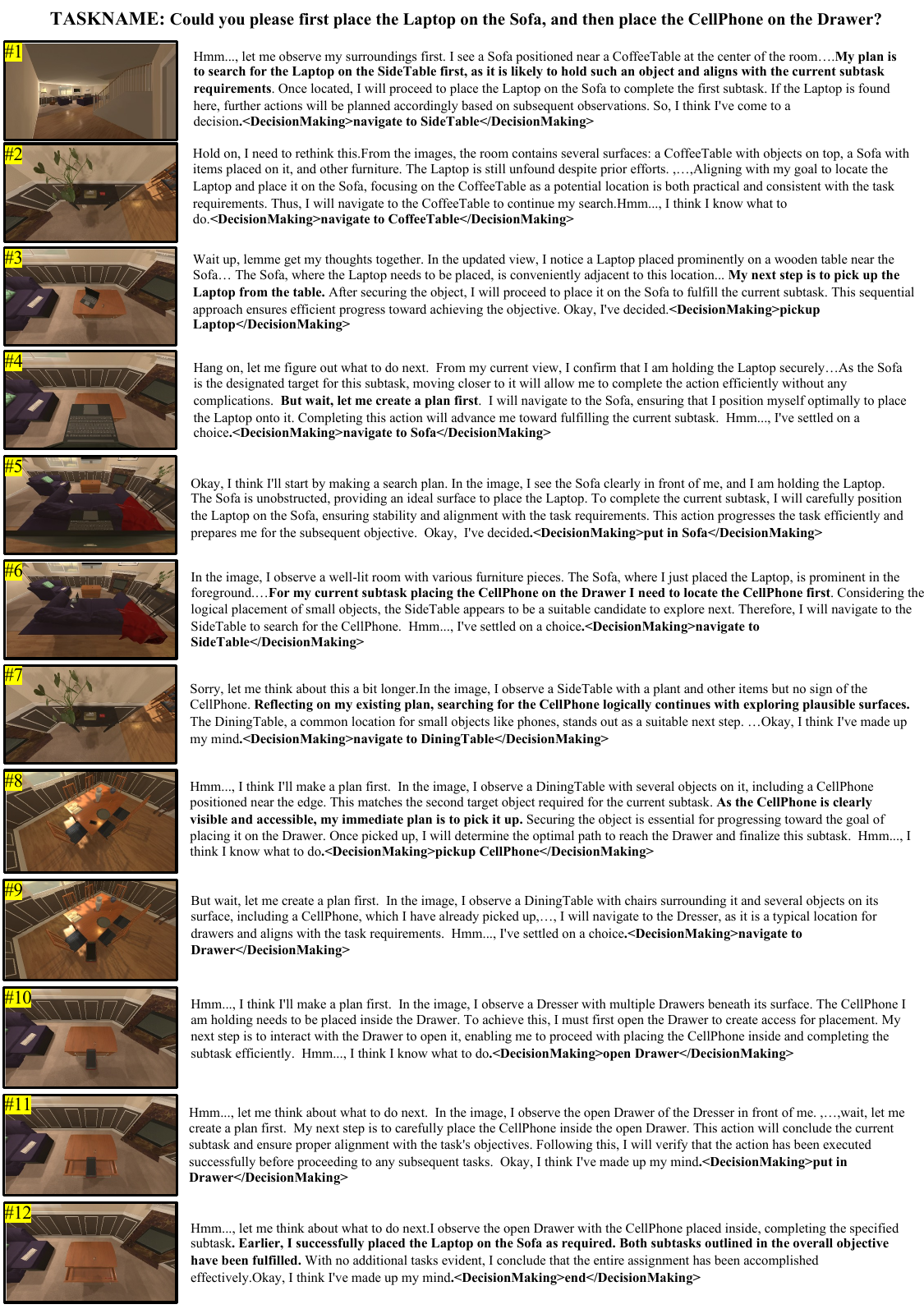} 
  \caption{Trajectory Case for Embodied Reasoner}
  \label{case_our1}
\end{figure*}

As shown in~\Cref{case_our1}, it illustrates Embodied-Reasoner’s performance on the task: "Could you please first place the Laptop on the Sofa, and then place the CellPhone on the Drawer?"

\subsection{GPT-o1's Action Trajectories}\label{gpt-o1}
\begin{figure*}[t!]
  \centering
   \includegraphics[width=0.9\linewidth]{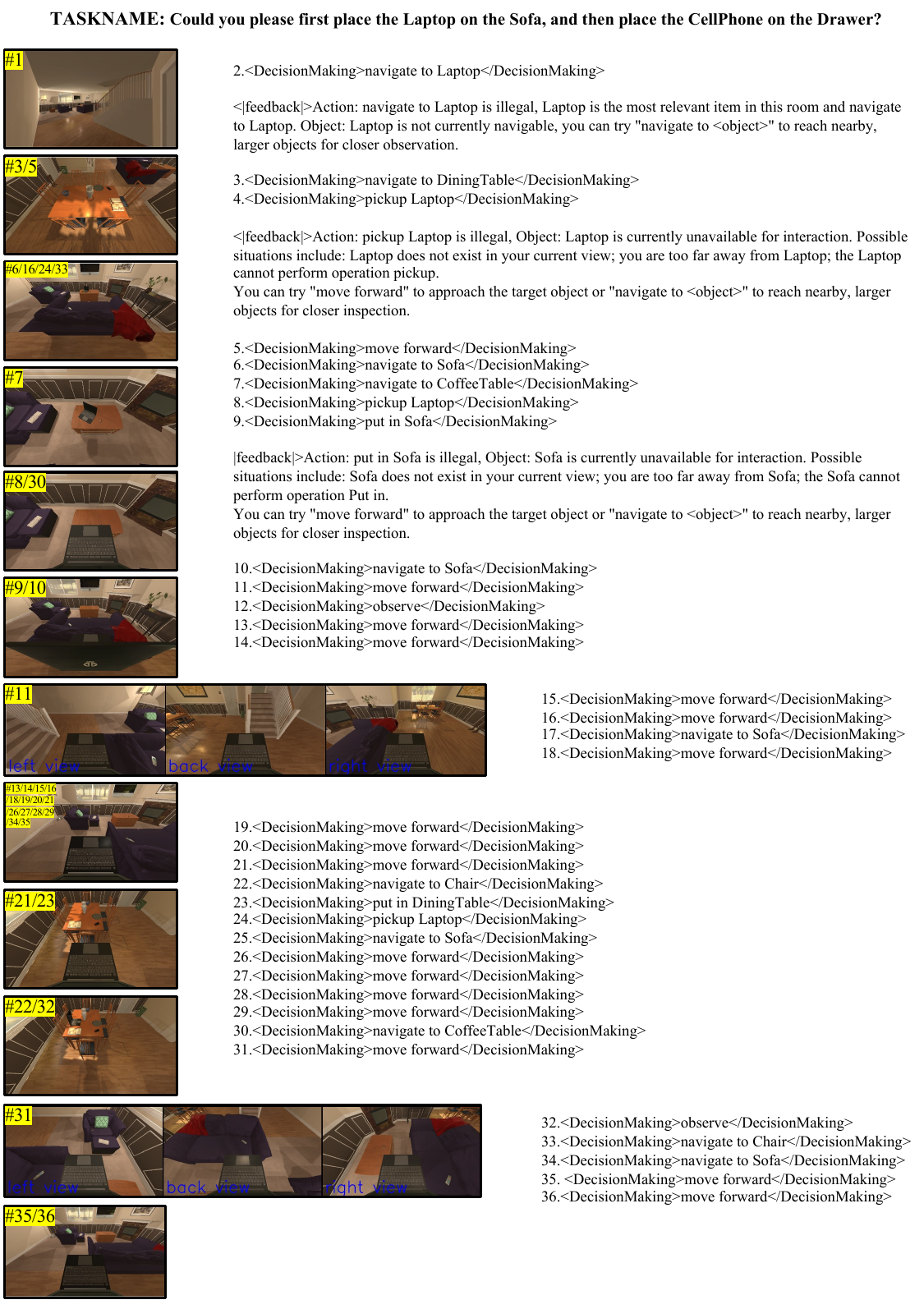}
   \caption{Trajectory Case for GPT-o1}
   \label{o1}
\end{figure*}

As shown in~\Cref{o1}, it illustrates GPT-o1’s performance on the task: "Could you please first place the Laptop on the Sofa, and then place the CellPhone on the Drawer?"

In comparison, during task trajectory, GPT-o1 sometimes forgets the task objective after navigating to the target container. For instance, after the 10th interaction, it navigates to the sofa but forgets to place the Laptop. Additionally, it occasionally falls into action loops, such as the repeated move forward actions from steps 13 to 16, 18 to 21.

In contrast, Embodied-Reasoner performs reasoning and planning, searches multiple locations, and retains memory of the current step’s objective, ultimately completing the task successfully.

\subsection{Embodied-Reasoner in Real-World}\label{Embodied-Reasoner-real-world}
\begin{figure*}[t!]
  \centering
   \includegraphics[width=0.95\linewidth]{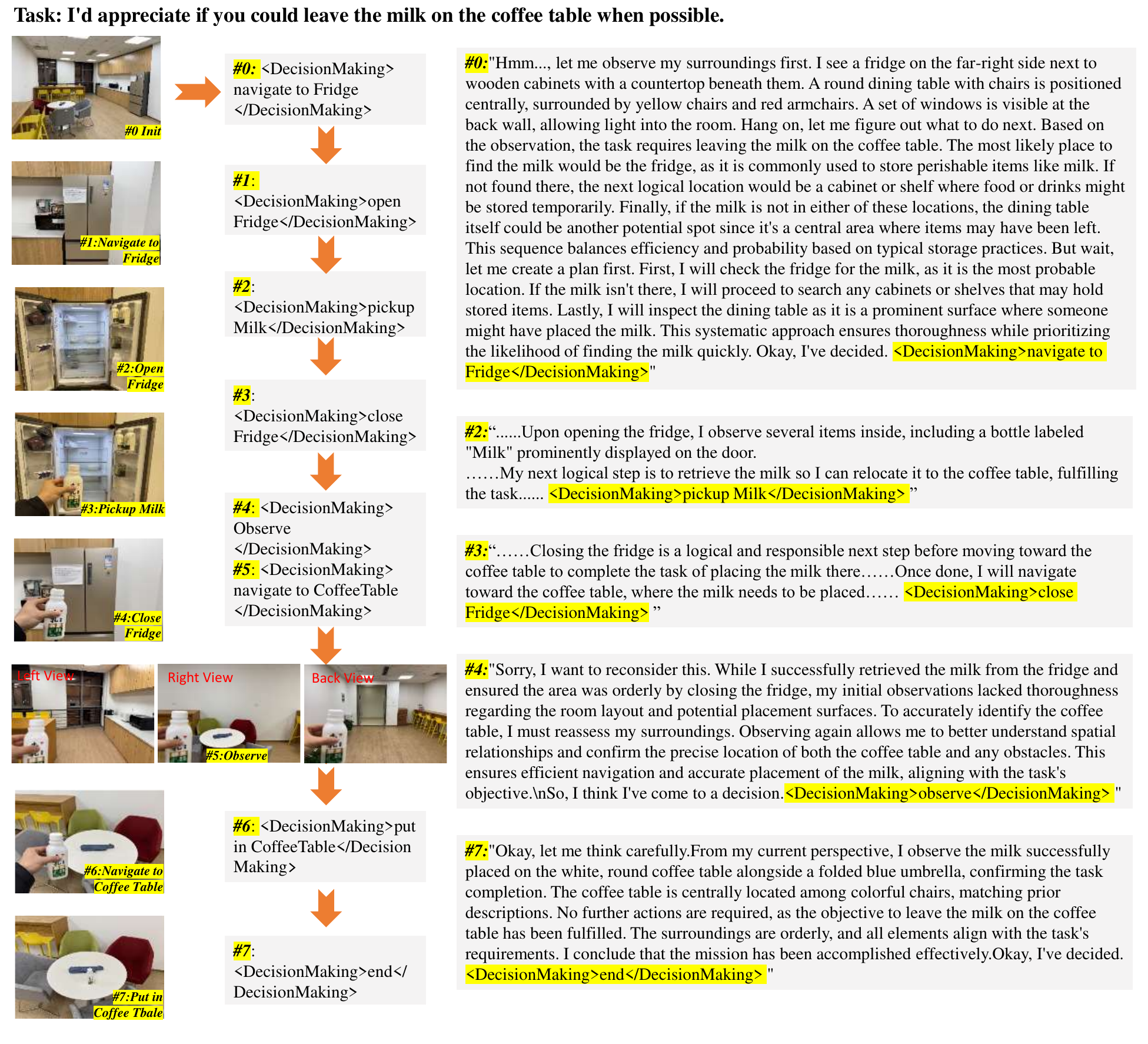}
   \caption{Trajectory Case for Embodied Reasoner in Real World}
   \label{case_our_real_world}
\end{figure*}

As shown in~\Cref{case_our_real_world}, it illustrates Embodied-Reasoner’s performance in the real world for the task: "I’d appreciate it if you could leave the milk on the coffee table when possible". 
It shows that Embodied-Reasoner can complete the task in real-world scenarios through step-by-step planning and reasoning.

\end{document}